\documentclass[10pt,twocolumn,letterpaper]{article}

\usepackage{cvpr}              

\usepackage{float}
\usepackage{newfloat}

\usepackage{algorithm}
\usepackage{algorithmic}
\usepackage{amsmath}
\usepackage{amssymb}
\usepackage{booktabs}
\usepackage{subcaption}

\definecolor{cvprblue}{rgb}{0.21,0.49,0.74}
\usepackage[pagebackref,breaklinks,colorlinks,allcolors=cvprblue]{hyperref}

\def\paperID{*****} 
\def\confName{CVPR}
\def\confYear{2026}

\title{NAMI: Efficient Image Generation via Bridged Progressive Rectified Flow Transformers}

\author{
Yuhang Ma$^{1}$\footnotemark[1] ~~
Bo Cheng$^{1}$\footnotemark[1] ~~
Shanyuan Liu$^{1}$\footnotemark[1] ~~
Hongyi Zhou$^{2}$\footnotemark[1] ~~
Liebucha Wu$^{1}$\\
Dawei Leng$^{1}$\footnotemark[2] ~~
Yuhui Yin$^{1}$ \\ 
$^1$360 AI Research \quad $^2$Tsinghua University
}

\begin{document}
\twocolumn[{%
\maketitle
\begin{figure}[H]
    \hsize=\textwidth
    \centering
    \includegraphics[width=1.0\textwidth]{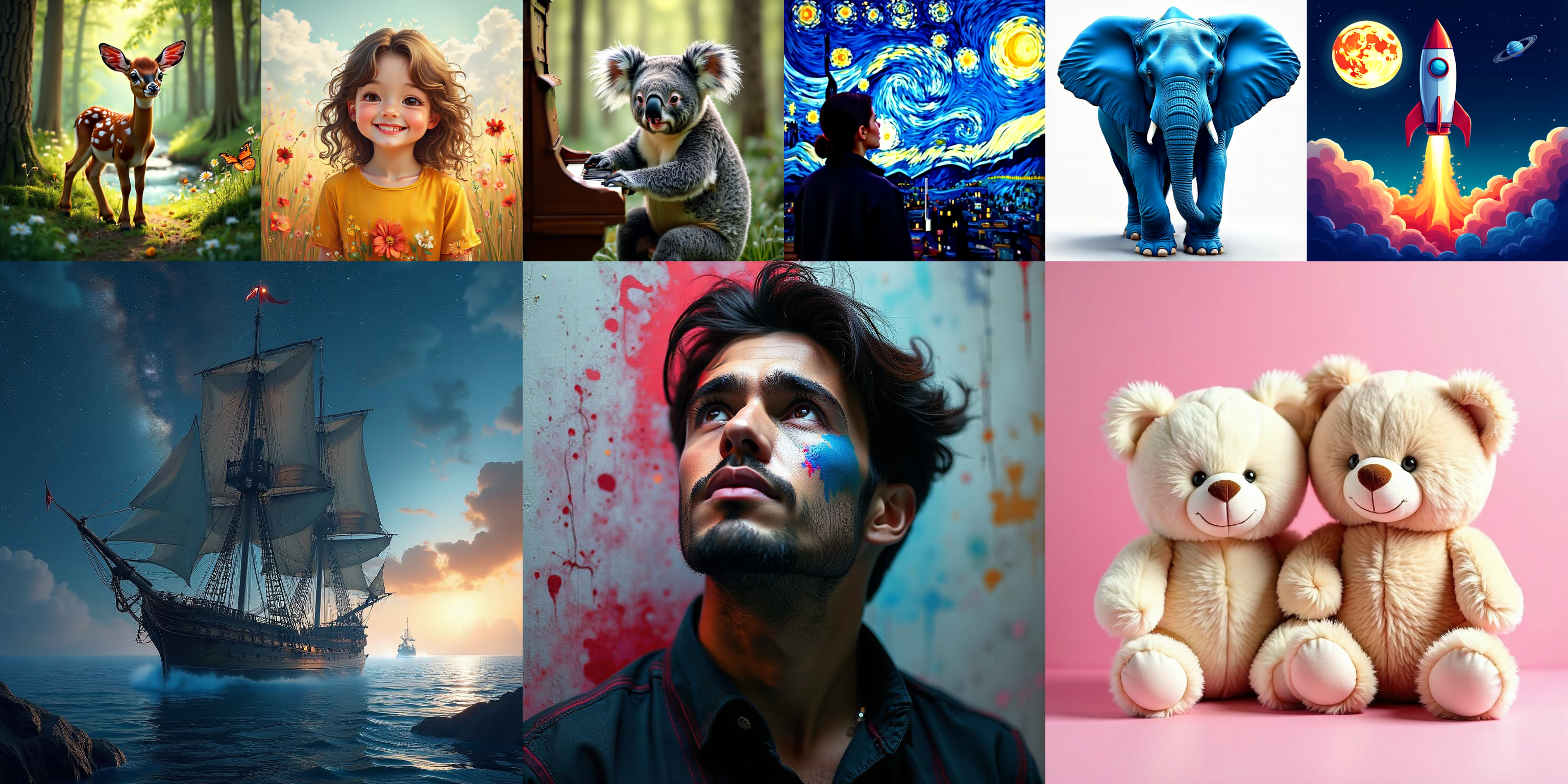}
        
    \captionof{figure}{\label{fig:1} High-quality image synthesis results from NAMI-2B demonstrate its capabilities in precise prompt following, spatial reasoning, and aesthetic quality.}
\end{figure}
}]
{
  \renewcommand{\thefootnote}%
    {\fnsymbol{footnote}}
  \footnotetext[1]{Equal contribution. \quad mayuhang@360.cn}
  \footnotetext[2]{Corresponding author. \quad lengdawei@360.cn}
}
\begin{abstract}
Flow-based Transformer models have achieved state-of-the-art image generation performance, but often suffer from high inference latency and computational cost due to their large parameter sizes. To improve inference efficiency without compromising quality, we propose Bridged Progressive Rectified Flow Transformers (NAMI), which decompose the generation process across temporal, spatial, and architectural demensions. We divide the rectified flow into different stages according to resolution, and use a BridgeFlow module to connect them. Fewer Transformer layers are used at low-resolution stages to generate image layouts and concept contours, and more layers are progressively added as the resolution increases. Experiments demonstrate that our approach achieves fast convergence and reduces inference time while ensuring generation quality. The main contributions of this paper are summarized as follows: (1) We introduce Bridged Progressive Rectified Flow Transformers that enable multi-resolution training, accelerating model convergence; (2) NAMI leverages piecewise flow and spatial cascading of Diffusion Transformer (DiT) to rapidly generate images, reducing inference time by 64\% for generating 1024×1024 resolution images; (3) We propose a BridgeFlow module to align flows between different stages; (4) We propose the NAMI-1K benchmark to evaluate human preference performance, aiming to mitigate distributional bias and comprehensively assess model effectiveness. The results show that our model is competitive with state-of-the-art models.
\end{abstract}    
\section{Introduction}
\label{sec:introduction}
Over the past year, text-to-image (T2I) models based on both diffusion~\cite{sohl2015deep,ho2020denoising,rombach2022high,dhariwal2021diffusion} and autoregressive approaches~\cite{chen2020generative,ramesh2021zero} have achieved significant advancements in image generation quality and inference efficiency. The diffusion model approach, represented by SD3~\cite{esser2024scaling} and FLUX~\cite{blackforestlabs2024}, leverages rectified flow~\cite{liu2022flow,albergo2022building,lipman2022flow} and MM-DiT~\cite{esser2024scaling} architectures to achieve excellent performance. However, the increased model parameters have significantly raised training and inference costs, making commercialization more challenging. Concurrently, autoregressive (AR) models generate images through next-token prediction~\cite{parmar2018image}. These methods use vector quantization variational autoencoder (VQVAE) technology~\cite{van2017neural} to discretize image tokens, and then predict the next component of the image based on the previously generated tokens. The size of the codebook plays a critical role in determining both the efficiency of image generation and the image quality. Currently, AR models such as LlamaGen~\cite{sun2024autoregressive} and Infinity~\cite{han2025infinity} exhibit significant advantages in inference speed, but their image generation quality still falls short.
\begin{figure}[H]
  \centering
  
   \includegraphics[width=1.0\linewidth]{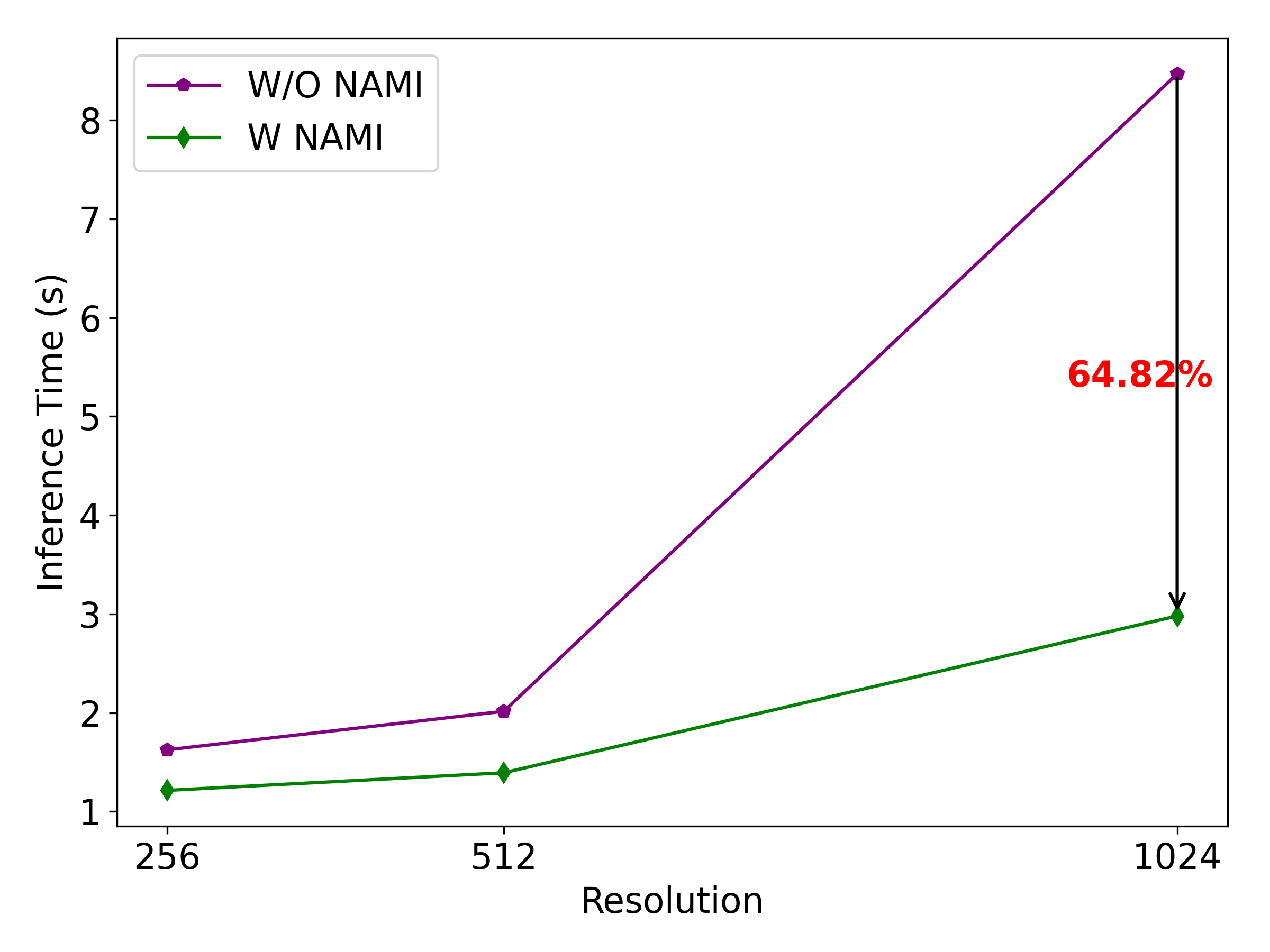}
   
   \caption{An overview of inference latency between the proposed NAMI-2B and the corresponding FLUX-2B base model of the same size without NAMI. With NAMI, inference performance improvement becomes more significant as image resolution increases. The measurements are conducted with a batch size of 1 on an A100 GPU.}
   \label{fig:2}
\end{figure}
Current research on optimizing the training and inference speed of diffusion models is primarily focused on areas such as latent space downsampling, reducing the number of tokens input into the DiT~\cite{peebles2023scalable} block, and efficient attention mechanisms~\cite{wang2024qihoo}. Several approaches have been proposed to reduce the number of input image tokens, one of which involves decomposing the image generation process from low resolution to high resolution. The matryoshka diffusion model (MatryoshkaDM)~\cite{gu2023matryoshka} innovatively integrates the low-resolution diffusion process as part of the high-resolution generation, utilizing a nested UNet~\cite{ronneberger2015u} architecture to accelerate model training. Another approach involves simultaneous upsampling of the image and denoising within the diffusion model, where the pyramid flow matching~\cite{jin2024pyramidal} method breaks down the video generation process into multiple resolution stages for independent denoising. While these methods improve the efficiency of image training or generation, each still has its own limitations.
\begin{figure}[H]
  \centering
  
   \includegraphics[width=1.0\linewidth]{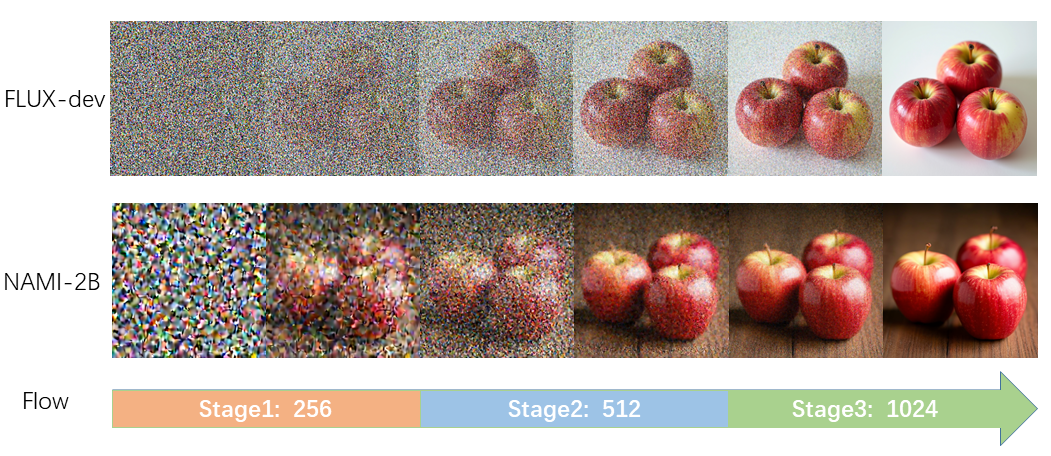}
   
   \caption{Overview of the image generation process for FLUX-dev~\cite{blackforestlabs2024} and our NAMI-2B, with upscaling alignment applied during the low-resolution stages of NAMI-2B.}
   \label{fig:progressive}
\end{figure}
We observe that in the image generation process of diffusion models, rough conceptual placements and outline layouts are performed in the early stages, while detail enhancement occurs in the later stages. Currently, most methods perform unified denoising across all sampling stages without considering the underlying mechanics of the image generation process. The early layout generation process can not only be quickly executed at low resolution but can also be modeled using a subset of the model parameters. Therefore, jointly considering temporal segmentation in the sampling strategy and the rational allocation of the model space is crucial to improving the efficiency of image generation.

In this paper, we propose a spatiotemporal separation progressive framework, NAMI. We divide the rectified flow into different stages based on resolution, connecting stages with BridgeFlow. Fewer transformer layers are used at the low-resolution stages to generate image layouts and concept contours, while more layers are progressively added as the resolution increases. As shown in the \cref{fig:progressive}, we visualize the image generation process of FLUX-dev~\cite{blackforestlabs2024} and NAMI-2B under the same inference steps. Despite using a lower resolution and fewer model parameters in the early stages, NAMI-2B still effectively generates image layouts and concepts. Furthermore, considering the overall process, NAMI-2B achieves a faster generation speed.

Current benchmarks such as GenEval~\cite{ghosh2023geneval}, DPG-Benchmark~\cite{hu2024ella}, and ImageReward~\cite{xu2023imagereward}, are primarily used to evaluate a model's performance. However, these benchmarks suffer from issues such as limited caption quantity, distribution biases. Therefore, a more comprehensive benchmark is essential for accurately evaluating a model's capabilities. We combine open benchmarks, AI-generated, and human-created test prompts to construct a multidimensional evaluation dataset named NAMI-1K.

In summary, Our primary contributions are as follows:

\begin{enumerate}
    \item Our method enables direct training on multi-resolution images, facilitating rapid learning of semantics at low resolutions and obtain details and quality at high resolutions, thereby accelerating model convergence.
    \item The NAMI-2B employs temporal separation and spatial cascading within the model to efficiently generate images. As shown in \cref{fig:2}, using the same model size, our approach reduces the time required to generate 1024 resolution images by 64\%, while still maintaining a high level of image quality. 
    \item Furthermore, we propose a BridgeFlow module to align flows between different stages.
    \item We propose the NAMI-1K benchmark and evaluate our model on both NAMI-1K and open-source benchmarks. The results across multiple metrics consistently highlight the effectiveness of the proposed method.
\end{enumerate}
\section{Related Work}
\label{sec:related work}

Diffusion models~\cite{sohl2015deep,ho2020denoising,rombach2022high,dhariwal2021diffusion} have become a powerful framework in image generation, with latent diffusion models further enhancing both quality and efficiency. DiT~\cite{peebles2023scalable} has introduced a transformative change by replacing the conventional UNet~\cite{ronneberger2015u} architecture used in models like SDXL~\cite{podell2023sdxl} with a transformer~\cite{vaswani2017attention}-based framework. This advancement has led to the development of models such as PixArt-$\alpha$~\cite{chen2023pixart}, Hunyuan-DiT~\cite{li2024hunyuan}, LUMINA-Next~\cite{zhuo2024lumina}, SD3~\cite{esser2024scaling}, and FLUX~\cite{blackforestlabs2024}, which efficiently integrate multimodal information to improve text-to-image generation, thereby demonstrating the effectiveness of transformer-based approaches. Notably, SD3 and FLUX employ rectified flow~\cite{liu2022flow,albergo2022building,lipman2022flow}, which connects data and noise on a straight line, improving both training and inference efficiency. Additionally, the MM-DiT~\cite{esser2024scaling} block has been proposed, achieving state-of-the-art generation performance. Our work builds upon MM-DiT as the fundamental module while adopting rectified flow as a specific forward path selection strategy.

Some studies have focused on optimizing the performance of diffusion models. SANA~\cite{xie2024sana} utilizes a higher downsampling rate for VAE~\cite{chen2024deep} and replaces vanilla quadratic attention modules with linear attention to enhance efficiency. MicroDiT~\cite{sehwag2025stretching} reduces training costs by using masks to decrease the number of input tokens. CLEAR~\cite{liu2024clear} applies convolution-like local attention strategies, constraining interactions to a local window around the query token to achieve linear complexity. Although these methods provide efficiency improvements, they inevitably result in some degradation of image generation quality due to the high compression ratio of VAE~\cite{chen2024deep} or the reduction in token interactions. Our approach differs from these methods by decomposing the generation process, reducing model redundancy while preserving generation quality. Furthermore, our optimization strategy is orthogonal to the aforementioned methods, allowing for parallel integration.

The denoising process~\cite{ho2020denoising} in diffusion models can be viewed as an image generation process that progresses from coarse-grained layouts to fine-grained details. Some studies have modeled this process in conjunction with multi-scale image generation. For example, LAPGAN~\cite{denton2015deep} generates images by first producing low-resolution images and then feeding them into a high-resolution model. Pyramidal denoising diffusion probabilistic models~\cite{ryu2022pyramidal} have also adopted a similar approach. Pyramid flow matching~\cite{jin2024pyramidal} accelerates inference performance by implementing a pyramidal strategy over time steps. MatryoshkaDM~\cite{gu2023matryoshka} uses different-sized models at different resolution scales, enabling the training of diffusion models in pixel space. These methods have reduced training complexity and accelerated inference by leveraging multiscale resolutions. However, they have not fully addressed the issue of parameter redundancy in DiT~\cite{peebles2023scalable} models. Our work leverages models of varying sizes across different scale stages to maximize performance efficiency while maintaining generation quality.

\begin{figure*}[htbp]
  \centering
  
   \includegraphics[width=\textwidth]{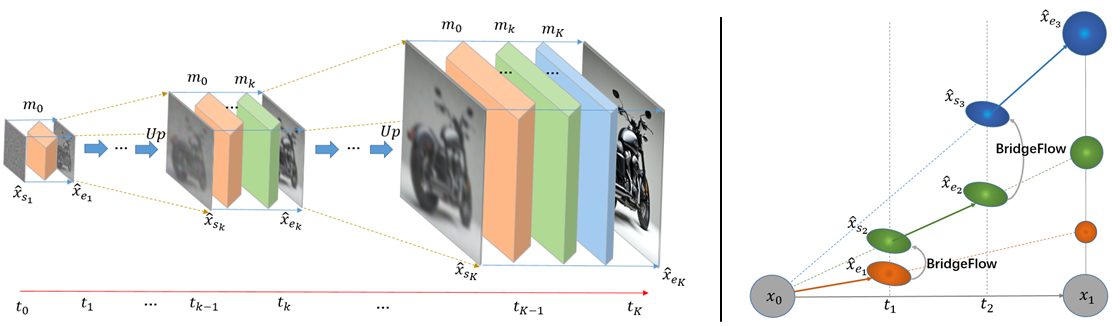}

   \caption{Overview of NAMI: The left figure shows the progressive flow transformers of NAMI, where the same color represents the same module. The right figure depicts the integration of the BridgeFlow module, which establishes connections across adjacent time windows. Specifically, we divide the image generation process into \( K \) resolution stages and the entire flow is divided into \( K \) time windows, where adjacent stages are connected through upsampling and the BridgeFlow module. We use fewer transformer layers at the low-resolution stages to generate image layouts and concept contours, progressively adding more layers as the resolution increases.}
   \label{fig:3}
\end{figure*}
\section{Method}
\label{sec:method}

\subsection{Preliminary Study}
Flow-based~\cite{papamakarios2021normalizing,xu2022poisson} generative models are similar to diffusion models, where the objective is to learn a velocity field \( v_\theta(x_t, t) \) that maps random noise \( x_0 \sim \mathcal{N}(0, I) \) to data samples \( x_1 \sim D \), via an ordinary differential equation (ODE):
\begin{equation}
  \frac{dx_t}{dt} = v_\theta(x_t, t), \quad x_0 \sim \mathcal{N}(0, I)
  \label{eq:1}
\end{equation}
Subsequently, a simple simulation-free training objective for flow generative models has been proposed. A representative method is rectified flow~\cite{liu2022flow,albergo2022building,lipman2022flow}, which adopts linear interpolation between the noise distribution \(x_0\)  and the data distribution \(x_1\). It trains a neural network to approximate the velocity field via the conditional flow matching loss. The optimization procedure is termed reflow:
\begin{equation}
\resizebox{0.9\linewidth}{!}{$\min_\theta \mathbb{E}_{x_0 \sim \mathcal{N}(0, I), x_1 \sim D} \left[ \int_0^1 \left\| (x_0 - x_1) - v_\theta(x_t, t) \right\|^2 \, dt \right]$}
  \label{eq:2}
\end{equation}
where \(\quad x_t = t x_1 + (1 - t) x_0 \).

\subsection{Progressive Rectified Flow Transformers}
\label{sec3.2}
Our method treats the image generation process as a transformation from a low-resolution image containing coarse-grained concepts to a high-resolution image with rich details. At low resolution, we employ piecewise flow~\cite{yan2024perflow} and partial modules to reduce computational costs. As shown in \cref{fig:3} , we divide the image generation into $K$ resolution stages, where $Up(.,.)$ denotes the upsampling function and $Down(.,.)$ denotes the downsampling function. For flow processing, we refer to pyramid flow matching~\cite{jin2024pyramidal} and divide the entire flow into $K$ time windows $ \{[t_{k-1},t_k]\}_{k=1}^K$, where $1=t_K > \dots > t_k > t_{k-1} > \dots > t_0=0$. Linear interpolation is also used between noise and data distribution, that is, $x_{t_k} = t_k x_1 + (1 - t_k) x_0$. The starting point for each time window is given by $\hat{x}_{s_k} = BridgeFlow(Up(Down(x_{t_{k-1}}, 2^{k+1})))$, which is obtained by upsampling the endpoint of the previous stage. The end point is given by $\hat{x}_{e_k} = Down(x_{t_k}, 2^k)$, which is obtained by downsampling \(x_1\) and adding noise. The flow within the window is given by:
\begin{equation}
\resizebox{0.9\linewidth}{!}{$
    \hat{x}_t = t' \hat{x}_{e_k} + (1 - t') \hat{x}_{s_k}, \quad t' = (t - t_{k-1})/(t_k-t_{k-1}). $}
    \label{eq:3}
\end{equation}
Additionally, we divide the corresponding model into $K$ modules, where each module $m_k$ ($k \in [1, K]$) consists of MM-DiT~\cite{esser2024scaling} blocks. For the $k$-th time window, the model is denoted as $\theta_k = \{m_1 \oplus \dots \oplus m_k\}_{k=1}^K$, where $\oplus$ represents the concatenation of the blocks. Finally, our overall optimization process can be expressed as:
\begin{equation}
\resizebox{0.9\linewidth}{!}{$
  \min_{\theta_k} \sum_{k=1}^K E_{(k,t,(\hat{x}_{s_k}, \hat{x}_{e_k}))} \left[ \int_{t_{k-1}}^{t_k} \left\| (\hat{x}_{s_k} - \hat{x}_{e_k}) - v_{\theta_k} (\hat{x}_t, t) \right\|^2 \, dt \right]$}
  \label{eq:4}
\end{equation}

\subsection{BridgeFlow Module}
To ensure the continuity of probabilistic paths across different stages, Pyramid Flow~\cite{jin2024pyramidal} treats the jump points as Gaussian distribution matches and derives the corresponding transitions of the mean and covariance matrices. During inference, it performs rescaling and renoising to handle these transitions. However, this process is non-parametric and lacks learnable adaptation, which limits its robustness. Moreover, the renoising step has a time complexity proportional to the image token length, resulting in unsatisfactory efficiency and performance.

To overcome these limitations, we propose the BridgeFlow module, which introduces a learnable linear transformation to align the probability distributions at stage boundaries in a data-driven manner. Specifically, for the endpoint of each stage, we first apply upsampling to match the resolution, followed by a linear transformation  $\hat{x}_{s_k} = W\cdot\mathrm{Up}(\hat{x}_{e_{k-1}})+B$ to match the distribution of the subsequent stage's starting point. Further analysis is provided in the Appendix C. In practice, every BridgeFlow module is pretrained with a mean squared error (MSE) loss according to the defined flow partitions. The detailed ablation results are presented in ~\cref{tab:jump_point}.

\subsection{Multi-Resolution Training}
\label{sec3.3}
\begin{algorithm}[ht]
\caption{Multi-resolution Training with Progressive Rectified Flow Transformers}
\label{algorithm:1}
\begin{algorithmic}[1]
\STATE \textbf{Input:} Number of windows $K$, Multi-resolution Datasets $D = \{D_k\}_{k=1}^{K}$ 
\STATE Create $K$ time windows $[t_{k-1}, t_k]$ for $k=1$ to $K$ with $t_K = 1, t_0 = 0$
\STATE Initialize $\theta_k = \{m_1 \oplus \dots \oplus m_k\}_{k=1}^{K}$
\REPEAT
    
    \FOR{each time window $[t_{k-1}, t_k]$}
        
        \STATE Sample $x'_1 \sim D'\{D_k, D_{k+1}, ..., D_K\}$
        \STATE do $Down(D',.)\ \text{until}\ x'_1 \sim \{D_k, D_k,...,D_k\} $
        \STATE Compute the start point 
        
        \IF{$k = 1$}
            \STATE $\hat{x}_{s_k} \gets x_0$
        \ELSE
            \STATE $\hat{x}_{s_k} \gets \text{BridgeFlow}(\text{Up}(\text{Down}(x'_{t_{k-1}}, 2)))$
        \ENDIF
        \STATE Compute the end point $\hat{x}_{e_k} = x'_{t_k}$
        
        \STATE Sample $t \in [t_{k-1}, t_k]$, compute $\hat{x}_t = t' \hat{x}_{e_k} + (1 - t') \hat{x}_{s_k}$ with $t'= (t - t_{k - 1}) / (t_k - t_{k - 1})$
        \STATE Compute the loss: 
        \[
        \text{loss} = \|(\hat{x}_{s_k} - \hat{x}_{e_k}) - v_{\theta_k} (\hat{x}_t, t)\|^2
        \]
    \ENDFOR
    \STATE Weighting the losses for different time windows and perform backpropagation
\UNTIL{convergence}
\end{algorithmic}
\end{algorithm}

Considering knowledge sharing between modules and efficient training, we propose a multi-resolution progressive approach during the training process. Different resolutions of data can be used at different stages, and the optimization follows the approach in \cref{eq:4}. The training process is shown in Algorithm \ref{algorithm:1}. 
Unlike the conventional method of training at low resolutions first and then fine-tuning at higher resolutions separately, our approach enables the joint training of data with different resolutions simultaneously. This facilitates knowledge sharing within the model and helps prevent catastrophic forgetting during fine-tuning at high resolutions. At each stage, we train using images with resolutions greater than or equal to the current resolution. The \(Down(.,.)\) function is applied to downsample the images to the corresponding resolution, from which the starting and ending points are computed to obtain the target and loss for that stage. The loss of different stages is jointly optimized by dynamically adjusting their weights according to the training process. Additionally, during training, we also follow SD3~\cite{esser2024scaling} and use logit-normal sampling for each stage. To enable CFG~\cite{ho2022classifier}, we apply a 0.1 probability for random dropping of the prompt.

\subsection{Inference}

During inference, we sample a noise at the minimum resolution, start from stage $k=1$ and proceed until $k=K$, where at each stage, we use the Flow-Euler-discretization sampler. For the jump points between stages, We use upsampling and BridgeFlow to perform the transformation. Our approach can reduce 64\% time during inference.
\section{Experiments}
\label{sec:experiments}
\subsection{Model Details}
In this paper, we use the MM-DiT block from FLUX~\cite{blackforestlabs2024} to build our NAMI due to its state-of-the-art performance in text-to-image generation. As shown in \cref{tab:nami_configurations}, our NAMI consists of 22 layers, each with 2048 channels, and uses 16 attention heads, totaling 2B parameters. Additionally, we perform ablation experiments by scaling the model down to 0.6B parameters. The layer distribution for different stages of the two models is visible in the table under layers ratio. For the text encoder, we use mT5~\cite{xue2021mt5} and mCLIP~\cite{chen2023mclip} to enable our model with multilingual capabilities.

\begin{table}[ht]
  \centering
  
    \caption{Architecture details of the proposed NAMI.}
    \label{tab:nami_configurations}
      \begin{tabular}{@{}lcccc@{}}
        \toprule
        Model & Width & Depth & Heads & Layers Ratio\\
        \midrule
        NAMI-0.6B & 1536 & 12 & 12 & 5:4:3 \\
        NAMI-2B   & 2048 & 22 & 16 & 9:7:6 \\
        \bottomrule
      \end{tabular}
  
\end{table}

\subsection{Inference Time Details}

As shown in \cref{tab:nami_inference}, we provide the inference time of the NAMI-2B model at each resolution, using 10 inference steps per stage. The "Overall" time additionally includes the upsampling and BridgeFlow modules at the connection points between stages. Compared to the Flux-based 2B model with the same total of 30 sampling steps, our inference time is significantly reduced 64\%.
The design of the flow piece according to resolution reduces the computation time by 53\%, while the model partition further decreases the time by an additional 11\%. Refer to \cref{fig:inference_time} for details.
\begin{table}[ht]
  \centering
  
  \caption{The inference time details of the NAMI architecture (measured in seconds).}
  \label{tab:nami_inference}
    \begin{tabular}{@{}lccccc@{}}
      \toprule
      
      Method & 256 & 512 & 1024 & Overall & Reduction \\
      \midrule
      Baseline & - & - & 8.47 & 8.47 & - \\
      NAMI-2B & 0.27 & 0.45 & 2.21 & 2.98 & \textbf{{64.82\%}} \\
      \bottomrule
    \end{tabular}
  
\end{table}

\subsection{Experiments Details}
We trained our NAMI-2B model on LAION~\cite{schuhmann2022laion}, GRIT-20M~\cite{peng2023kosmos}, further fine-tuned it with 100K high-quality internal data, applying semantic and aesthetic filtering to all datasets. The total size of the training set is approximately 100 million. We set the generated image resolution to 1024 and the number of stages $K$ to 3. The resolution for each stage is set to 256, 512, and 1024, respectively, with the corresponding time window division ratio being 1:1:1.

We first pretrain the BridgeFlow modules between stages according to the time window division, using a learning rate of 1e-3. Convergence is typically reached within 10k training steps. Subsequently, we trained the NAMI-2B model using the multi-resolution simultaneous training strategy described in Method. At the early stage of training, we set the learning rate to 1e-4 and sampled three resolutions within each batch at a ratio of 4:2:1 (from low to high resolution) for 120k steps. Subsequently, we adjusted the sampling ratio to 2:4:1 and continued training for a further 80k steps. Finally, we modified the ratio to 1:2:4, reduced the learning rate to 5e-5, and trained for 50k steps to obtain the final NAMI-2B model.

\subsubsection{Quantitative Comparison}

We conducted a quantitative evaluation of NAMI across multiple open-source benchmarks, comparing our approach with other open-source methods.
\Cref{tab:geneval} demonstrates the text-image alignment capability for short prompts on GenEval~\cite{ghosh2023geneval}. It can be observed that our approach achieves a leading overall ranking when compared under comparable model parameter scales. \Cref{tab:dpg} shows the complex semantic alignment and instruction-following capability for long texts on DPG-Benchmarks~\cite{hu2024ella}. Our model still demonstrates exceptional performance in certain dimensions.
\begin{table}[htbp]
\centering
\caption{Comparison of different methods on GenEval. With highlight the \textbf{best}, \underline{second best} entries. Ovr \& Sgl \& Two \& Cnt \& Col \& Pos \& CA mean: Overall \& Single \& Two \& Counting \& Colors \& Position \& Color Attribution.}
\label{tab:geneval}
\setlength{\tabcolsep}{0.8mm}

\begin{tabular}{@{}lccccccc@{}}
\hline
Models(Params) & Ovr & Sgl & Two & Cnt & Col & Pos & CA \\
\hline
FLUX-dev(12B) & \underline{0.67} & \textbf{0.99} & \underline{0.81} & \textbf{0.79} & 0.74 & 0.20 & \underline{0.47} \\
FLUX-schnell(12B) & \textbf{0.71} & \textbf{0.99} & \textbf{0.92} & \underline{0.73} & 0.78 & \underline{0.28} & \textbf{0.54} \\
LUMINA-Next(2B) & 0.46 & 0.92 & 0.46 & 0.48 & 0.70 & 0.09 & 0.13 \\
SDXL(2.6B) & 0.55 & 0.98 & 0.74 & 0.39 & 0.85 & 0.15 & 0.23 \\
Hunyuan-DiT(1.5B) & 0.63 & 0.97 & 0.77 & 0.71 & \textbf{0.88} & 0.13 & 0.30 \\
SD3-medium(2B) & 0.62 & 0.98 & 0.74 & 0.63 & 0.67 & \textbf{0.34} & 0.36 \\
Sana(1.6B) & 0.66 & \textbf{0.99} & 0.77 & 0.62 & \textbf{0.88} & 0.21 & \underline{0.47} \\
\hline
NAMI-2B(2B) & 0.65 & \textbf{0.99} & 0.78 & 0.64 & 0.82 & 0.20 & 0.45 \\
\hline
\end{tabular}

\end{table}

We found that GenEval~\cite{ghosh2023geneval} and DPG-Benchmark~\cite{hu2024ella} primarily focus on text-image alignment, with limited prompt diversity. 

\begin{table}[htbp]
\centering
\caption{Comparison of different methods on DPG-Benchmark. With highlight the \textbf{best}, \underline{second best} entries. Ovr \& Gbl \& Ent \& Attr \& Rel \& Oth mean: Overall \& Global \& Entity \& Attribute \& Relation \& Other}
\label{tab:dpg}
\setlength{\tabcolsep}{1.0mm}

\begin{tabular}{@{}lcccccc@{}}
\hline
Models(Params) & Ovr & Gbl & Ent & Attr & Rel & Oth \\
\hline
FLUX-dev(12B) & 84.0 & 82.1 & 89.5 & 88.7 & \underline{91.1} & 89.4 \\
FLUX-schnell(12B) & \textbf{84.8} & \textbf{91.2} & \underline{91.3} & \textbf{89.7} & 86.5 & 87.0 \\
LUMINA-Next(2B) & 74.6 & 82.8 & 88.7 & 86.4 & 80.5 & 81.8 \\
SDXL(2.6B) & 74.7 & 83.3 & 82.4 & 80.9 & 86.8 & 80.4 \\
Hunyuan-DiT(1.5B) & 78.9 & 84.6 & 80.6 & 88.0 & 74.4 & 86.4 \\
SD3-medium(2B) & 84.1 & 87.9 & 91.0 & 88.8 & 80.7 & 88.7 \\
Sana(1.6B) & \textbf{84.8} & 86.0 & \textbf{91.5} & \underline{88.9} & \textbf{91.9} & \textbf{90.7} \\
\hline
NAMI-2B(2B) & 83.8 & \underline{90.3} & 89.4 & 88.3 & 88.0 & \underline{90.6} \\
\hline
\end{tabular}

\end{table}

\subsubsection{Human Evaluation}
To address the limited prompt diversity of open-source benchmarks and to accurately evaluate the model's performance, We propose the NAMI-1K evaluation dataset from the perspective of human preference performance.
\begin{table}[H]
  \centering
  \caption{Human evaluation results on NAMI-1K dataset. Rele \& Cohe \& Aes \& Real mean: Relevance \& Coherence \& Aesthetic \& Realism. With highlight the \textbf{best}, \underline{second best} entries.}
  \label{tab:human_evaluation_results}
  \setlength{\tabcolsep}{1.0mm}
  
  \begin{tabular}{@{}lccccc@{}}
    \toprule
    
    Models(Params) & Rele & Cohe & Aes & Real & Overall \\
    \midrule
    Flux-dev(12B) & \textbf{83.93} & \textbf{83.28} & \textbf{84.26} & \textbf{90.16} & \textbf{85.05} \\
    
    SD3-medium(2B) & 75.74 &65.90 &61.64 &75.74 &69.97 \\
    Infinity(2B) & \underline{76.39} &65.25 &61.97 &74.43 &69.77 \\
    Hunyuan-DiT(1.5B) & 74.43 & \underline{69.51} & \underline{63.93} &64.92 &68.95 \\
    SANA(1.6B) & 75.41 &62.30 &60.00 &72.46 &67.80 \\
    \hline
    NAMI-2B(2B) & 76.07 &66.89 &62.30 &\underline{76.72} &\underline{70.69} \\
    \bottomrule
  \end{tabular}
  
\end{table}

\textbf{Dataset Construction} We developed a hierarchical evaluation dataset, NAMI-1K, comprising 1,000 prompts with diverse topic categories and varying length distributions. This dataset was constructed by integrating open benchmarks, AI-generated prompts, and human-authored prompts. 
Specifically, 360 short prompts were selected from the open benchmarks GenEval~\cite{ghosh2023geneval} and LumiereSet~\cite{bar2024lumiere} to characterize the alignment capability of text to image. While 320 human-created prompts were collected from community contributions and user interactions to reflect the model's performance in real-world user scenarios. Additionally, 320 long prompts generated by Cogvlm2~\cite{hong2024cogvlm2}, were used to assess performance in complex semantic alignment and instruction-following capabilities. 

\begin{figure}[H]
  \centering
   \includegraphics[width=1.0\linewidth]{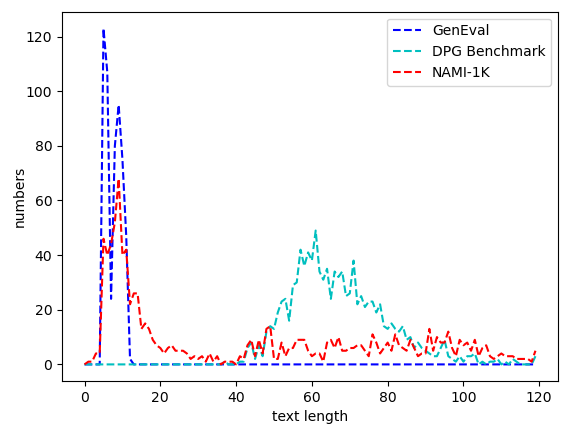}
   \caption{The distribution of text lengths across GenEval, DPG-Benchmark and NAMI-1K.}
   \label{fig:NAMI-1k-length}
\end{figure}

As shown in \cref{fig:NAMI-1k-length}, the prompt lengths in GenEval are primarily concentrated within 10 words, while those in the DPG-benchmark are mainly distributed between 50 and 80 words. In contrast, NAMI-1K exhibits a diverse distribution across different lengths within 120 words. As described in Appendix A, compared to GenEval and the DPG-benchmark, the topic distribution of NAMI-1K is more comprehensive and balanced. 

\textbf{Evaluation Process} The evaluation was conducted by five professional annotators through cross-assessment, considering four dimensions of relevance, coherence, aesthetic quality, and realism. The numerical range of each dimension is from 0 to 100, with a higher score indicating better performance. The final overall score was calculated through a weighted aggregation of the scores across the different dimensions, with the following weights assigned: relevance (0.3), coherence (0.3), aesthetic quality (0.2) and realism (0.2).

\textbf{Evaluation Results} The evaluation metrics are presented in \cref{tab:human_evaluation_results}. Although our model still trails behind FLUX-dev~\cite{blackforestlabs2024}, which have considerably larger parameter sizes, it shows a clear advantage over SD3-medium~\cite{esser2024scaling}, Infinity~\cite{han2025infinity}, Hunyuan-DiT~\cite{li2024hunyuan}, and SANA~\cite{xie2024sana}, which have comparable parameter sizes. As illustrated in Appendix F, a comparison of the generation results is provided between different methods.
\subsection{Ablation Study}
We sampled 2 million training data for the ablation experiment, and 1,000 for the test data. We used the NAMI-0.6B model for the experiments.

\textbf{Training Efficiency} We compare the training efficiency of NAMI with multi-resolution training against FLUX-based training of the same size at a final resolution of 256. The experiments are conducted using 8 A100 GPUs (80 GB memory each). As shown in \cref{tab:train efficiency} NAMI demonstrates higher training efficiency compared to the FLUX-based baseline.
\begin{table*}[t]
\centering
\caption{Comparison of Training Efficiency between NAMI and FLUX-based Architectures.}
\label{tab:train efficiency}
\setlength{\tabcolsep}{1.0mm}
\begin{tabular}{lcccccc}
\toprule
Model & Batch Size & Throughput(img/s) & 8xA100 GPU Hours & Peak Memory & FID & CLIP Score \\
\midrule
NAMI & 768 & 274 & 156 & 55G & 8.9293 & 25.57\\
FLUX-based & 768 & 241 & 176 & 67.6G & 9.76 & 25.20\\
\bottomrule
\end{tabular}
\end{table*}

\textbf{Components Effectiveness} We conducted ablation experiments to separately verify the effectiveness of flow piecewise and model partitioning. We conducted two sets of experiments at 256 and 512 resolutions, with the following configurations: (1) the conventional flow model of FLUX-base, (2) the semi-NAMI structure with only flow piecewise (W NAMI only flow), and (3) the full NAMI structure incorporating both flow piecewise and model spatial decomposition.

\begin{figure}[H]
    \centering
    
    \begin{subfigure}{0.49\columnwidth}  
    \includegraphics[width=\textwidth]{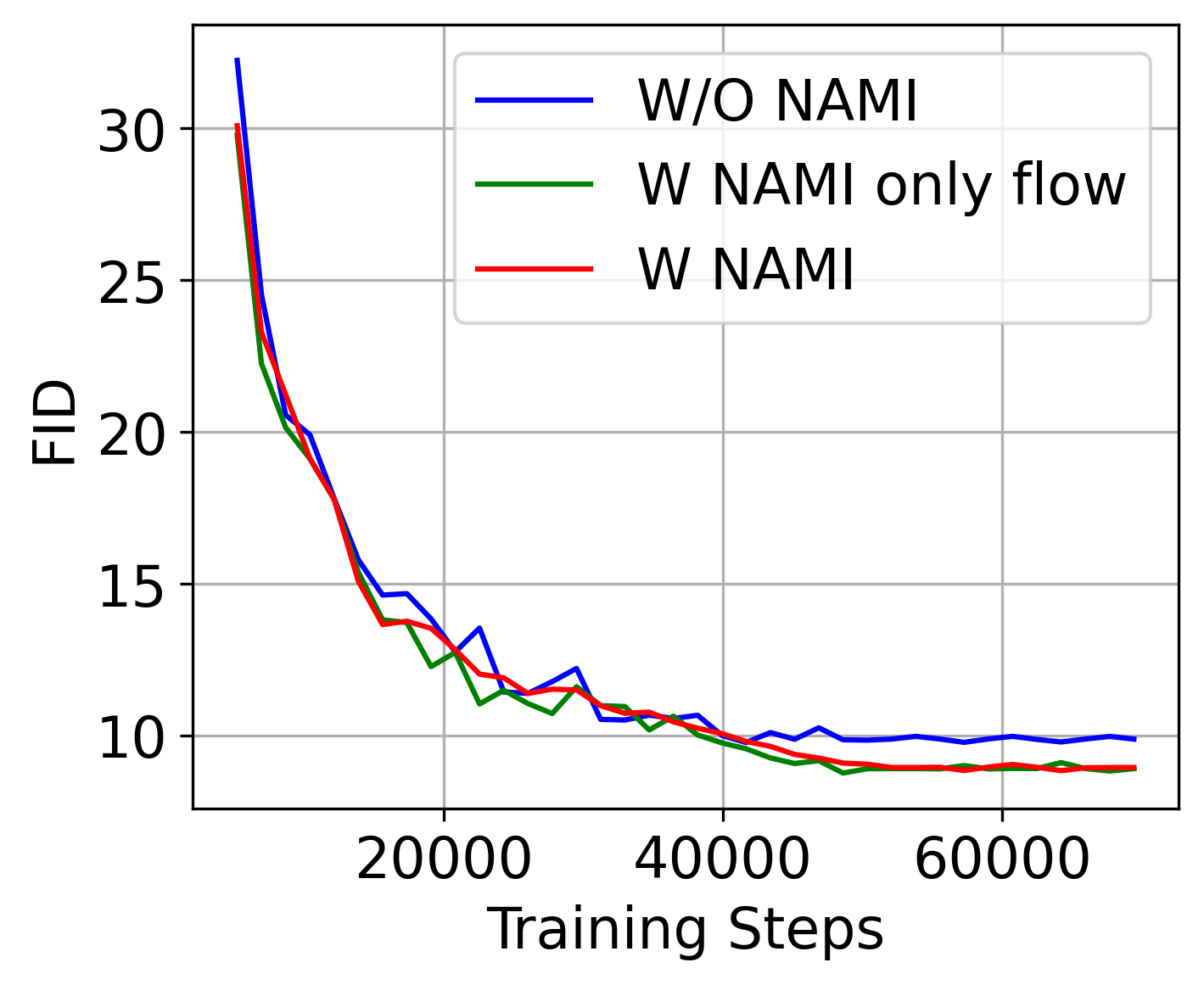}
    \caption{FID-256}
    \label{fig:effective_fid_256}
    
    \end{subfigure}\hfill  
    \begin{subfigure}{0.49\columnwidth}  
    \includegraphics[width=\textwidth]{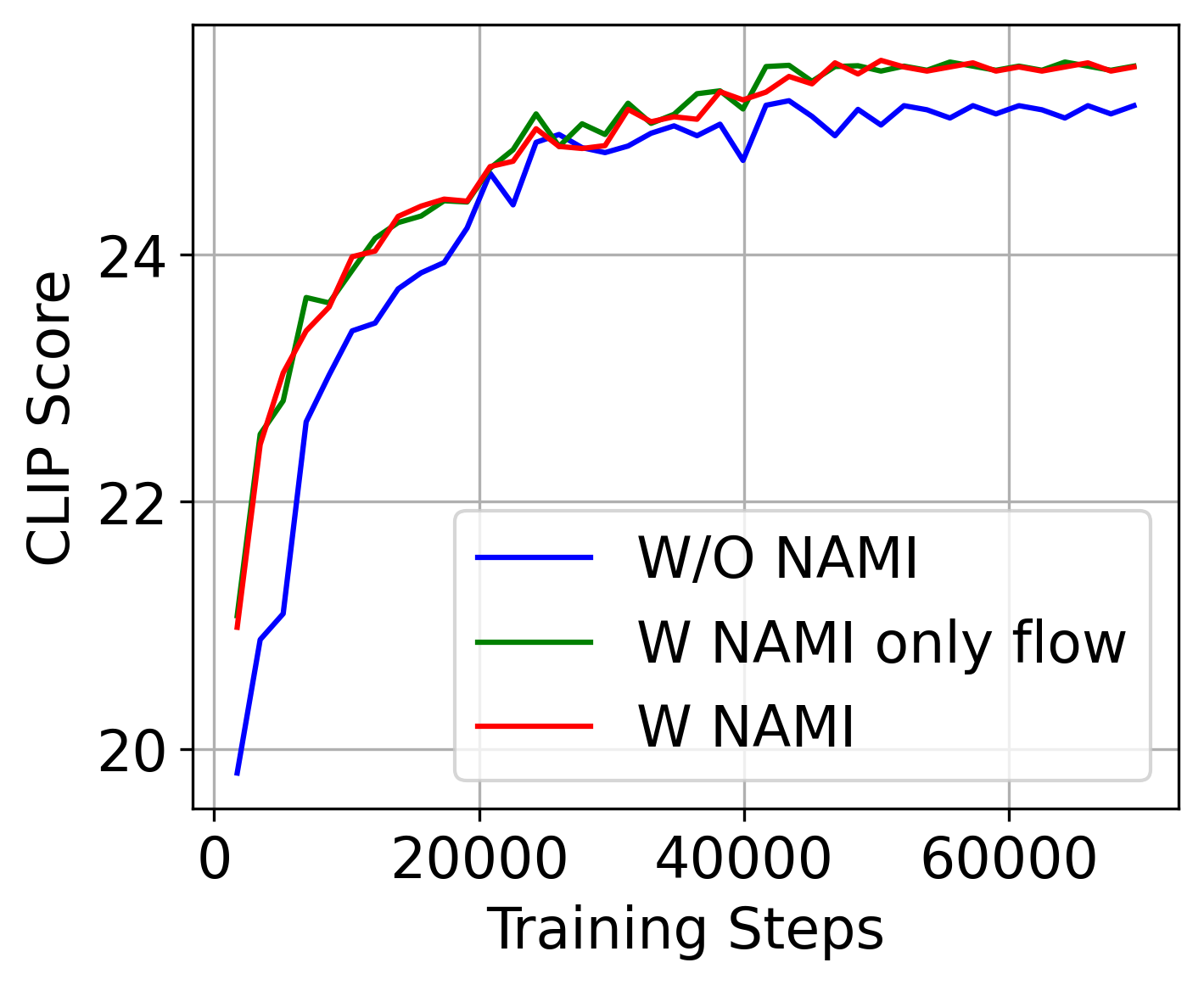}
    \caption{CLIP Score-256}
    \label{fig:effective_cs_256}
    \end{subfigure}
    \vfill
    
    \begin{subfigure}{0.49\columnwidth}  
    \includegraphics[width=\textwidth]{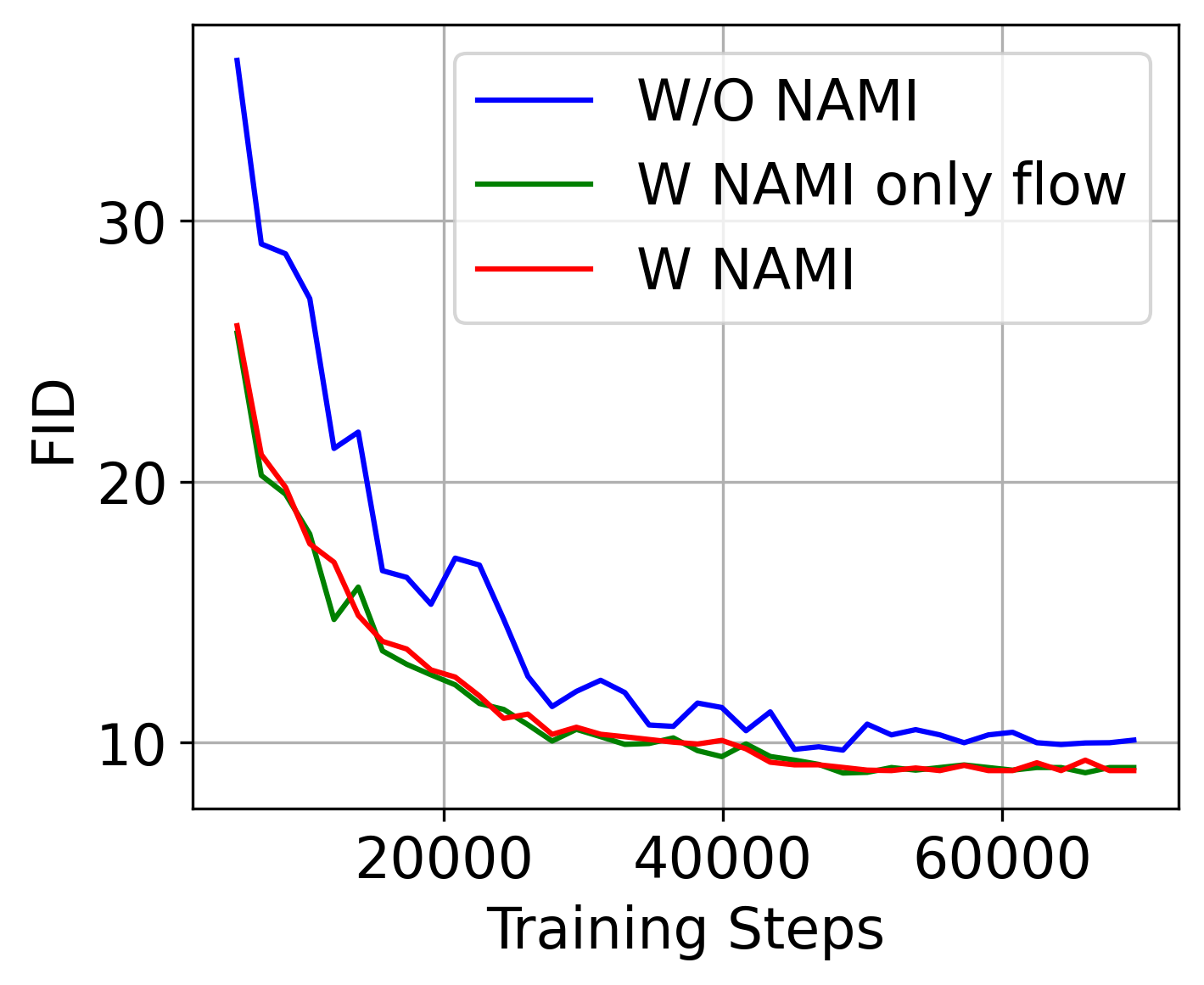} 
    \caption{FID-512}
    \label{fig:effective_fid_512}
    \end{subfigure}\hfill  
    \begin{subfigure}{0.49\columnwidth}  
    \includegraphics[width=\textwidth]{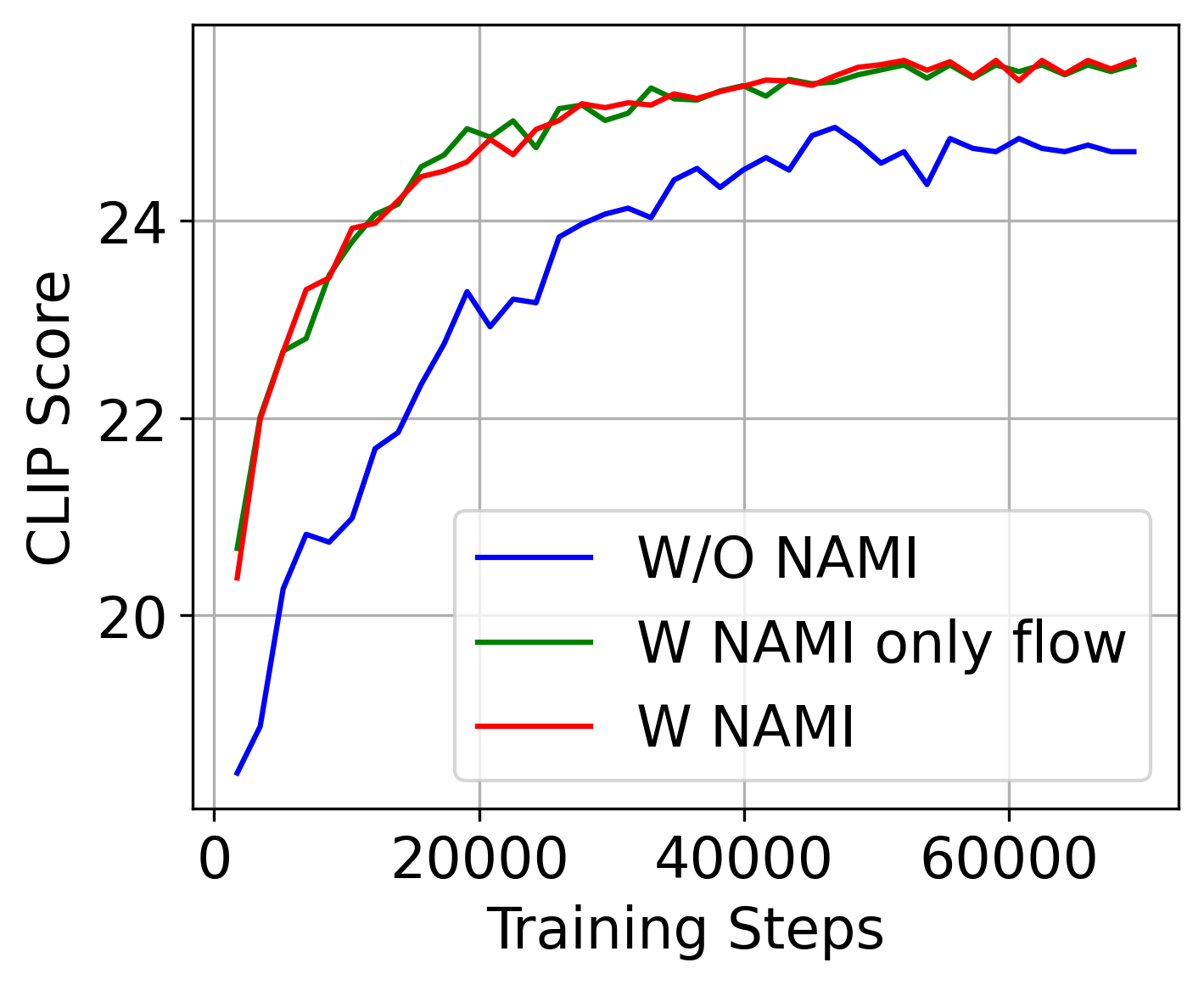} 
    \caption{CLIP Score-512}
    \label{fig:effective_cs_512}
    
    \end{subfigure}
  \caption{The effectiveness of the NAMI components at resolutions of 256 and 512.}
  \label{fig:effective_of_method}
\end{figure}

As shown in \cref{fig:effective_of_method}, experiment (1) and (3) show that NAMI converges faster, and achieves better FID~\cite{heusel2017gans} and CLIP~\cite{radford2021learning} scores at both resolutions. This advantage becomes more pronounced at the 512 resolution, further highlighting the effectiveness of the NAMI structure in accelerating high-resolution image training and improving generation quality. Experiment (1) and (2) demonstrate the effectiveness of piecewise flow and multi-resolution progressive generation. Experiment (2) and (3) compare the impact of model partitioning. We observe that although the full model has a slight convergence speed advantage in the early stages, the performance of both approaches is generally comparable. However, using partitioning provides a clear advantage in inference speed.
\textbf{Model Layers Partitioning} We conducted experiments on the model's layers ratio at a resolution of 512 to evaluate the impact of layers partitioning on performance. We adjusted the layer distribution ratio across three stages, with the total number of layers fixed at 12. 

\begin{figure}[H]
    \centering
    \begin{subfigure}{0.49\columnwidth}  
    \includegraphics[width=\textwidth]{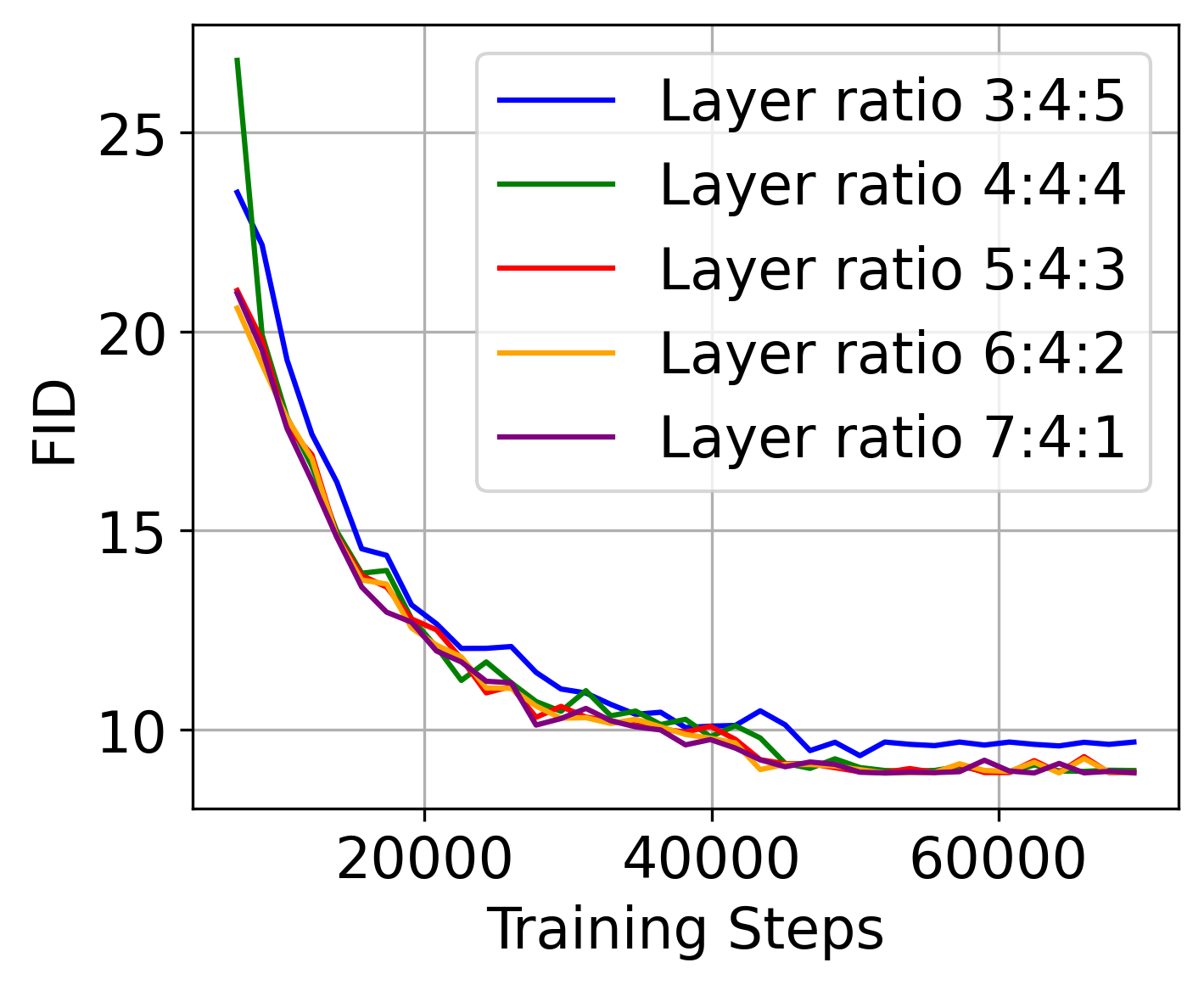}
    \caption{FID}
    \label{fig:layers_fid}
    \end{subfigure}\hfill  
    \begin{subfigure}{0.49\columnwidth}  
    \includegraphics[width=\textwidth]{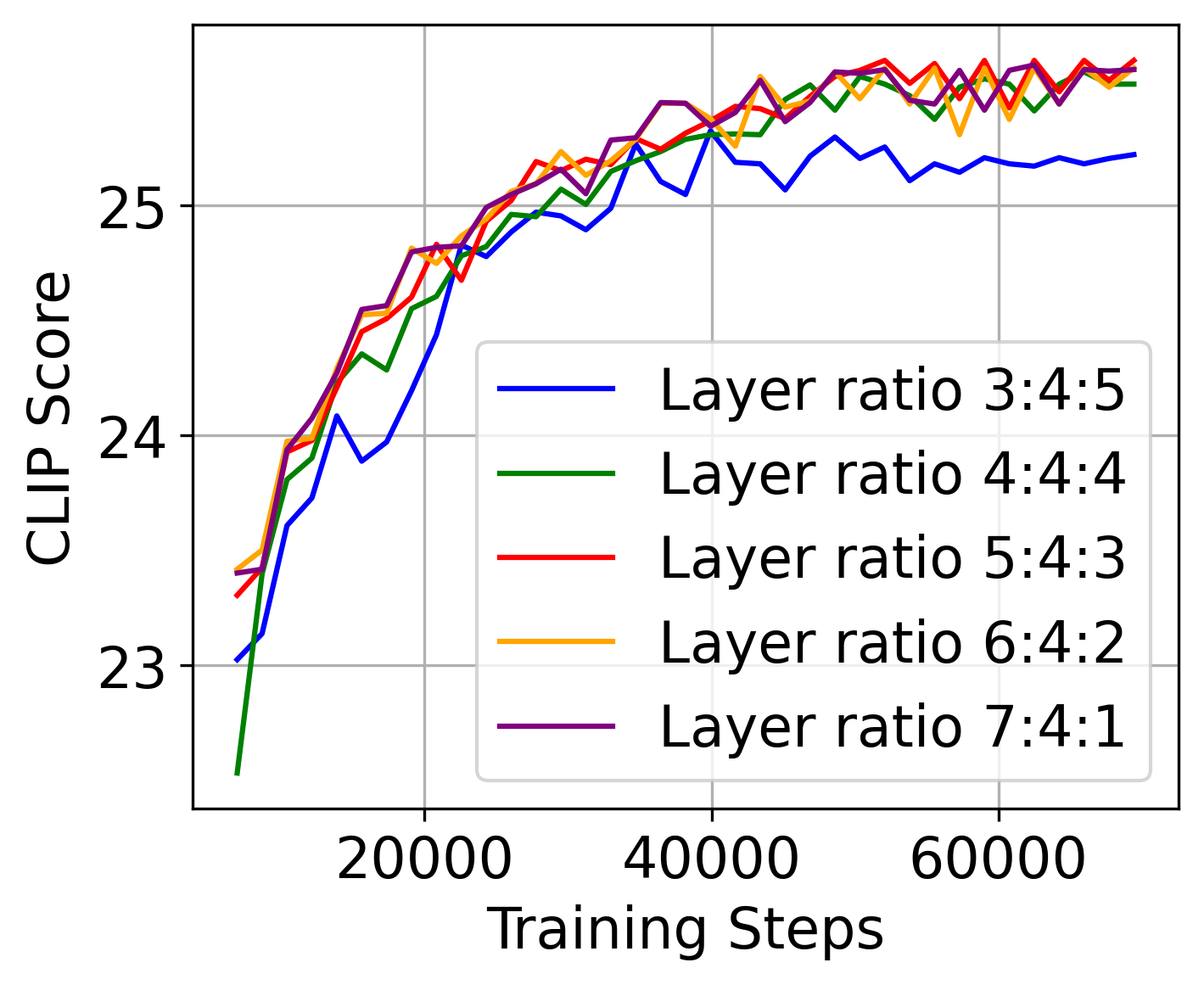}
    \caption{CLIP Score}
    \label{fig:layers_of_method}
    \end{subfigure}
  \caption{Ablation study on the distribution ratio of NAMI layers at different stages.}
  \label{fig:layers of method}
\end{figure}

As shown in \cref{fig:layers of method}, it can be seen that too few layers in the low-resolution stage lead to a decline in performance. However, as the number of layers increases to a certain point, the model's performance tends to saturate, and adding more layers results in redundancy in the model's capacity. In practical applications, this can be adjusted as a hyperparameter based on an equal distribution as a baseline.

\begin{figure}[H]
    \centering
    \begin{subfigure}{0.49\columnwidth}  
    \includegraphics[width=\textwidth]{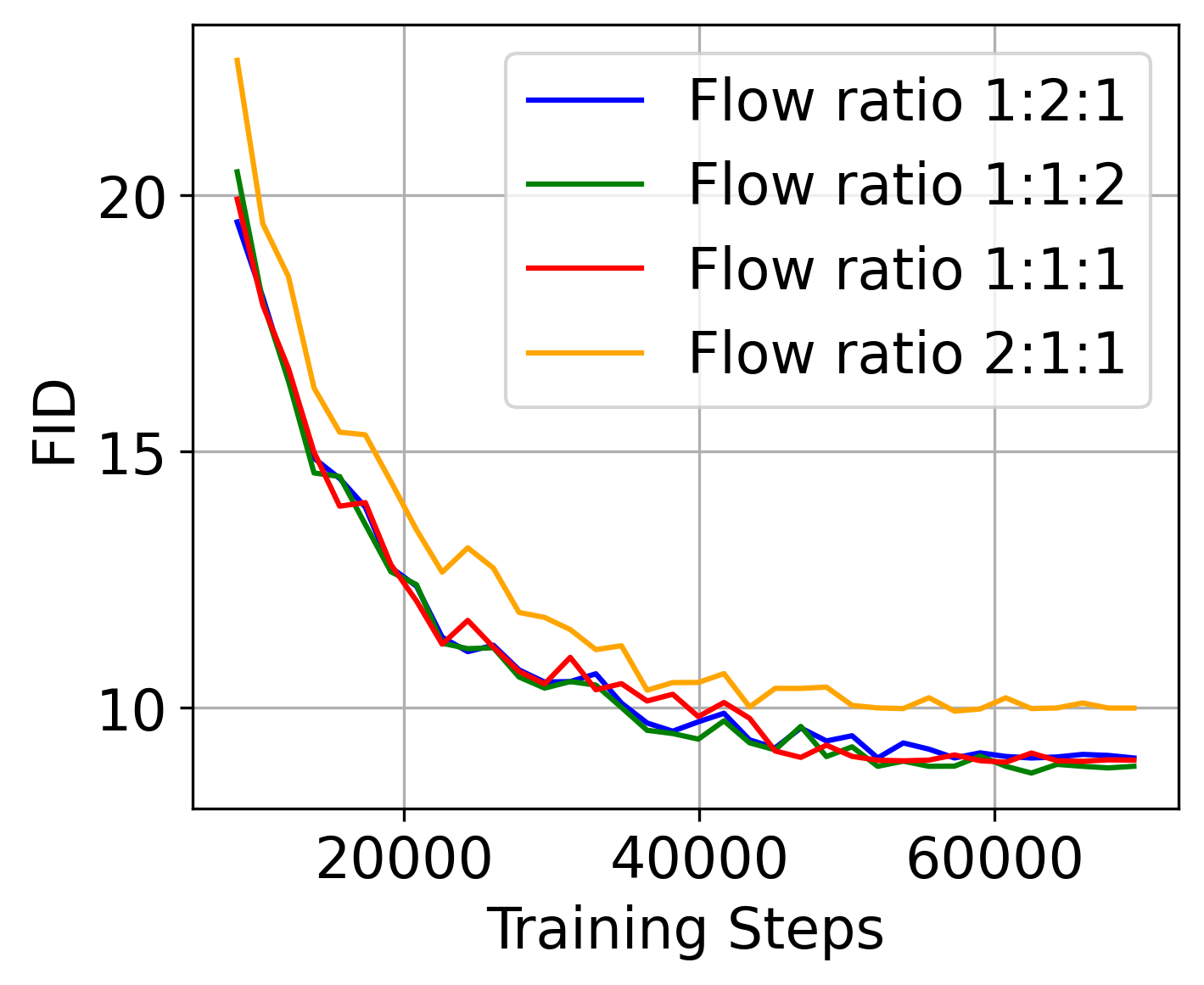}
    \caption{FID}
    \label{fig:flow fid}
    \end{subfigure}\hfill  
    \begin{subfigure}{0.49\columnwidth}  
    \includegraphics[width=\textwidth]{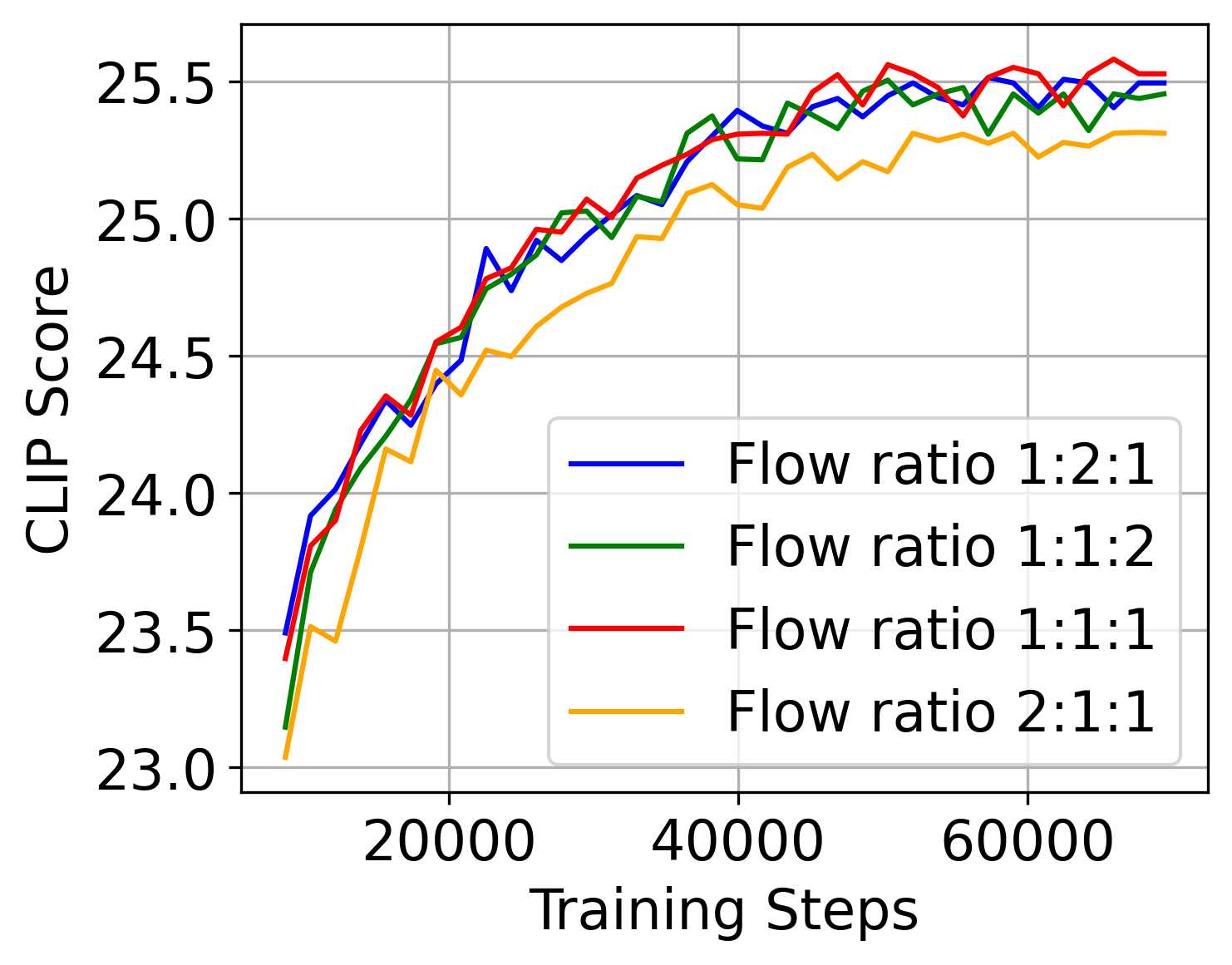}
    \caption{CLIP Score}
    \label{fig:flow cs}
    \end{subfigure}
  \caption{Ablation study on the flow piecewise ratio of NAMI at different stages.}
  \label{fig:flow of method}
\end{figure}

\textbf{Time Window Division Ratio} We adjusted the flow window distribution ratio across the three stages to observe the impact of different stage assignments. As shown in \cref{fig:flow of method}, it is evident that overly assigning the time windows to either the low or high stages does not result in significant improvements and may even lead to a decrease in accuracy due to insufficient length in other stages. Therefore, a uniform distribution of time windows is sufficient.

\textbf{BridgeFlow module} We conduct experiments at a resolution of 256 to compare different methods at the jump point, including: (1) Upsampling with Renoising as in pyramid flow matching~\cite{jin2024pyramidal}, (2) Upsampling followed by a 3-layer MLP, and (3) Pixel Shuffle~\cite{shi2016real} followed by a 3-layer residual CNN~\cite{he2016deep}. As shown in ~\cref{tab:jump_point}, PyramidFlow exhibits suboptimal performance, with its renoising process causing further increases in inference time at higher resolutions. Overall, our BridgeFlow module achieves a better trade-off between quality and speed, while introducing more complex structures does not yield further gains.

\begin{figure}[H]
  \centering
  
   \includegraphics[width=1.0\linewidth]{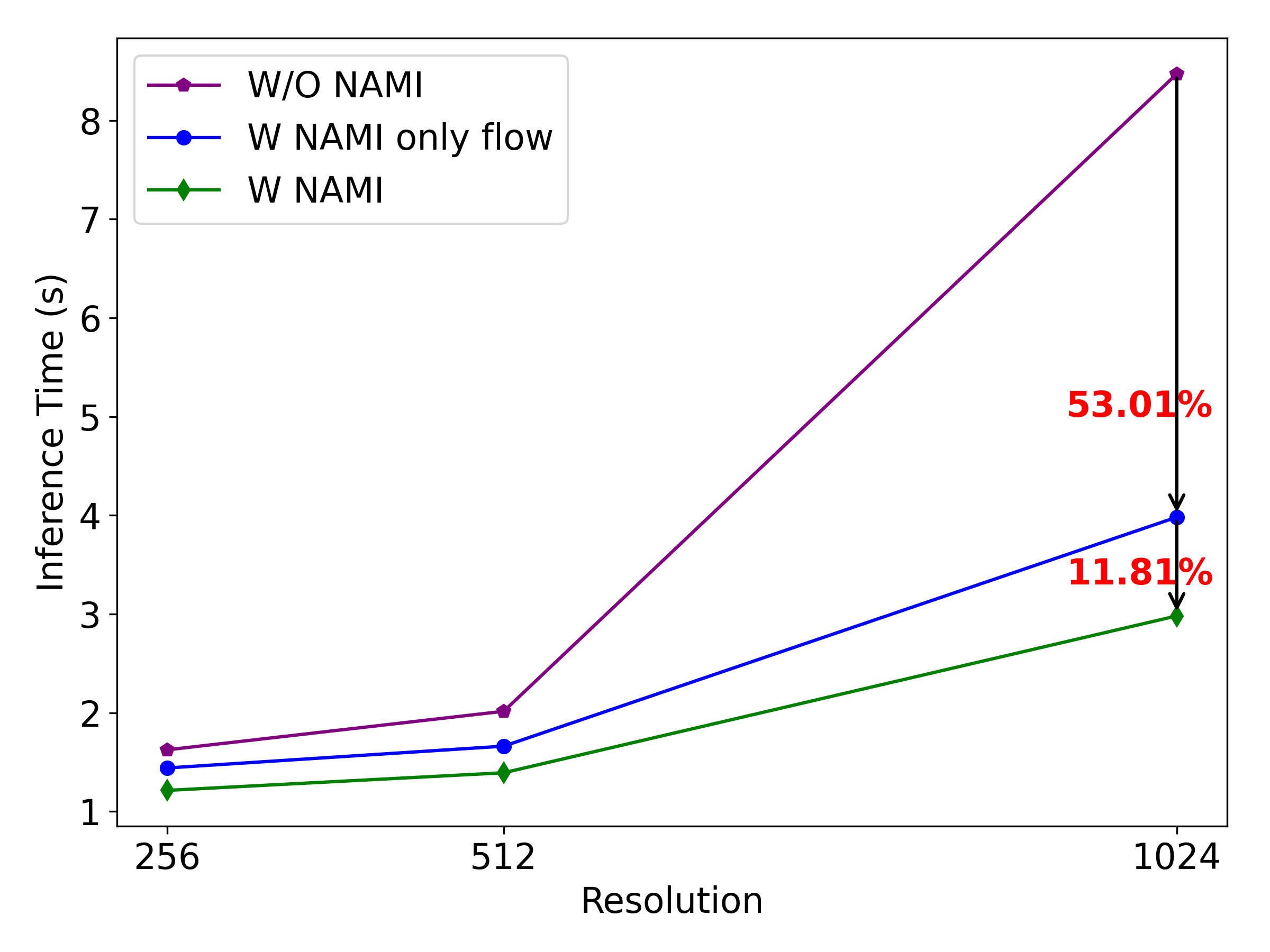}
   \caption{The inference time of NAMI Components.}
   \label{fig:inference_time}
\end{figure}

\begin{table}[t]
\centering
\caption{Comparison of different implementations at jump point for 256 resolution. Our proposed BridgeFlow module achieves the best overall performance, while more complex modules do not lead to further improvements.}
\label{tab:jump_point}
\setlength{\tabcolsep}{1.0mm}
\begin{tabular}{lccc}
\toprule
Module & Infer Time (s) & FID & CLIP Score \\
\midrule
BridgeFlow & 0.05 & 8.9293 & 25.57 \\
Upsample + MLP        & 0.16 & 9.0842 & 25.71 \\
Pixel Shuffle + CNN     & 0.23 & 8.8975 & 25.46 \\
Pyramid Flow                   & 0.12 & 9.8231 & 24.21 \\
\bottomrule
\end{tabular}
\end{table}
\textbf{The inference time of NAMI Components}
We performed 30 inference steps on an 80GB VRAM A100 GPU, excluding the time spent on the VAE and text encoders. We evaluated the inference time advantages brought by adjusting the flow and transformer components, respectively. As shown in  \cref{fig:inference_time}, NAMI method at a resolution of 1024 shows a 53.01\% reduction in inference time when only flow piecewise is applied. Additionally, with model partitioning, the inference time further decreased by 11.81\%.

\section{Conclusion}
\label{sec:conclusion}

In this paper, we introduce a bridge-connected  multi-stage flow and spatially nested model architecture and employs a multi-resolution joint training strategy to accelerate model convergence while ensuring performance. By dividing the rectified flow into multiple stages based on resolution and progressively increasing the number of transformer layers as the resolution grows, our method efficiently balances computational cost and performance. The proposed NAMI architecture reduces inference time by 64\% at 1024 resolution. Additionally, we have constructed the NAMI-1K benchmark to address the distribution biases and limited prompt diversity issues present in open-source benchmarks. NAMI is also convenient and promising for applications in other tasks. A simple training-free example of directly applying NAMI to image editing is provided in Appendix D. In future work, we plan to further explore NAMI’s potential in other tasks and its integration with other efficient methods. 

{
   \small
   \bibliographystyle{ieeenat_fullname}
   \bibliography{main}
}

\clearpage
\setcounter{page}{1}
\maketitlesupplementary

\section*{\textbf{A.} Detailed Description of NAMI-1K}
\label{A-NAMI-1k}
\begin{figure}[H]
  \centering
   
   \includegraphics[width=1.0\linewidth]{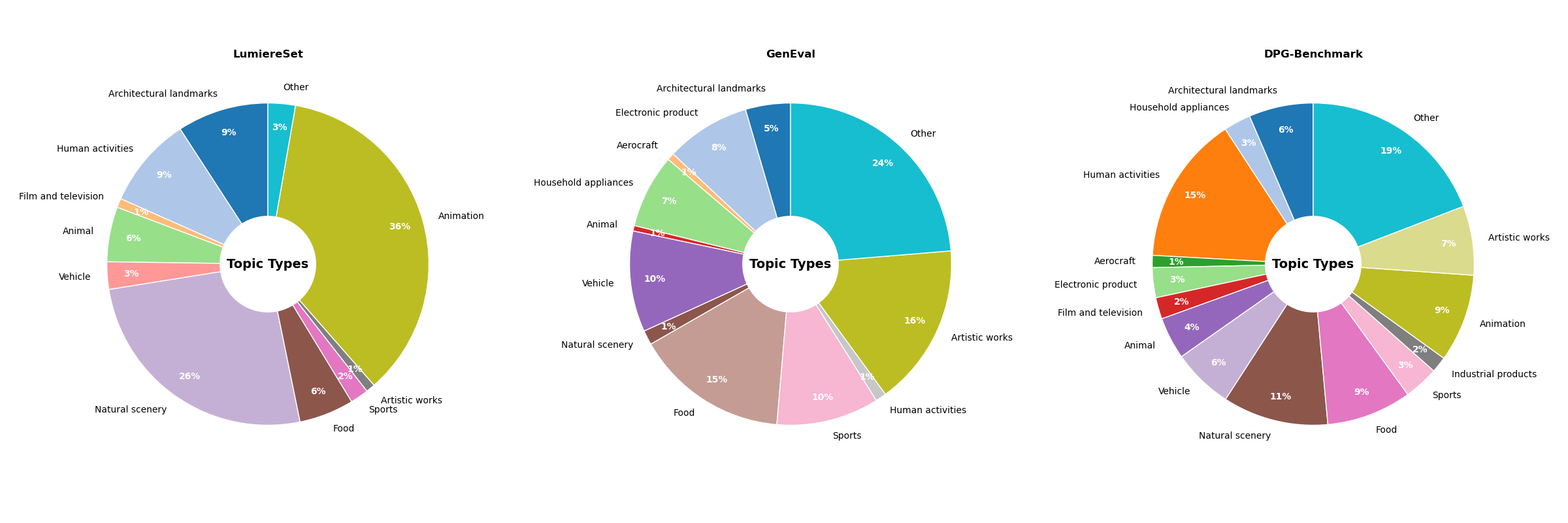}
   \caption{The distributions of topic types of GenEval, LumiereSet and DPG-Benchmark.}
   \label{fig:Other category}
\end{figure}

\begin{figure}[H]
  \centering
   
   \includegraphics[width=1.0\linewidth]{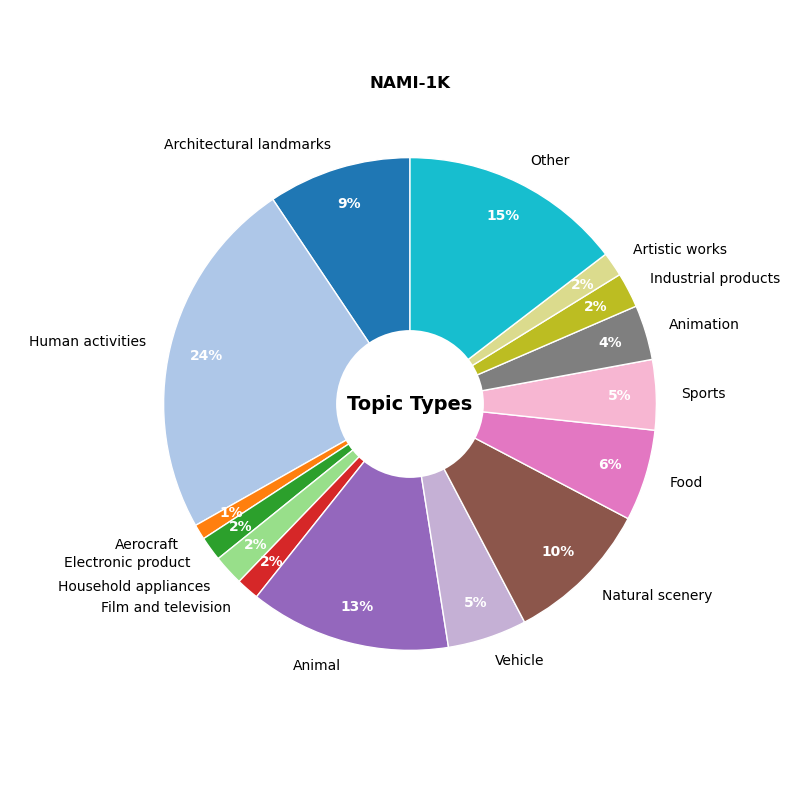}
   \caption{The distribution of topic types of NAMI-1K.}
   \label{fig:NAMI-1K category}
\end{figure}

NAMI-1K, consisting of 1,000 prompts  with diverse topic categories and varying length distributions. Specifically, 360 short prompts were selected from the open benchmarks GenEval and LumiereSet to characterize the alignment capability of text to image. While 320 human-created prompts were collected from community contributions and user interactions to reflect the model's performance in real-world user scenarios. Additionally, 320 long prompts generated by Cogvlm2, were used to assess performance in complex semantic alignment and instruction-following capabilities. 

As shown in Figure \ref{fig:Other category} and Figure \ref{fig:NAMI-1K category}, We selected the top 14 topics and analyzed the distribution of sample counts across these categories. Compared to GenEval and DPG-Benchmark, the NAMI-1K dataset exhibits a significantly richer variety of prompt topics and is more closely aligned with real-world application scenarios.

We select five prompts from each of the three major categories for demonstration, as outlined below.

\textbf{Open-benchmarks Prompts}

1. "A photo of four apples."

2. "A photo of a toothbrush and a carrot."

3. "A photo of a green bus and a purple microwave."

4. "A beautiful sunrise on mars, Curiosity rover. High definition, timelapse, dramatic colors."

5. "View of a castle with fantastically high towers reaching into the clouds in a hilly forest at dawn."

\textbf{Human-created Prompts}

1. "The Little Match Girl."

2. "A pair of sisters happily folding paper airplanes."

3. "A man hugging a tiger in a lucid nightmare."

4. "How to raise a healthy and happy dog."

5. "Chester Zoo staff member taking a picture of cheetah footprint."

\textbf{AI-generated Prompts}

1. "A man wearing a red hat, with a beard and mustache, and a red scarf around his neck. He is looking directly at the camera, giving the impression of a portrait. The man appears to be well-dressed, possibly in a suit, and his attire is complemented by the red hat and scarf. The overall atmosphere of the painting is one of sophistication and elegance."

2. "A blue gate with a white fence, adorned with pink flowers. The gate is open, allowing a view of the garden beyond. The garden is filled with various potted plants, some of which are placed on the ground, while others are positioned on the fence. The plants are of different sizes and colors, creating a vibrant and lively atmosphere. The combination of the blue gate, white fence, and the abundance of flowers and plants make the scene visually appealing and inviting."

3. "A colorful illustration of a rocket blasting off into space, with a bright moon in the background. The rocket is positioned towards the top of the scene, while the moon is located towards the left side. The illustration is set against a dark background, which contrasts with the vibrant colors of the rocket and the moon. The overall scene is visually appealing and captures the essence of space exploration."

4. "The afterglow of the setting sun casts a golden hue on the winding Great Wall, presenting a realistic scene. The bricks of the Great Wall appear ancient and sturdy under the golden sunlight, with each brick clearly visible. The distant mountains, illuminated by the sunset, show varying shades of orange and red, with distinct contours. Occasionally, a few birds fly over the Great Wall, adding a touch of vitality to the scene. In the background, the sky is dyed with brilliant shades of orange, red, and purple, with a few clouds scattered, making it exceptionally magnificent."

5. "A man in a black suit is standing on the street, holding a golden saxophone. His head is slightly bowed, and his eyes are closed, seemingly immersed in the music. The metallic sheen of the saxophone glistens in the sunlight, with details such as rings and keys clearly visible. The man's shoes are black leather, polished to a shine. The background features a busy city street, with a few cafes on the roadside, their tables and chairs neatly arranged outdoors, and several customers leisurely sipping coffee. In the distance, there are skyscrapers, with blue skies and white clouds highlighting the city's prosperity."

\section*{\textbf{B.} More Results of NAMI}

\begin{figure}[H]
  \centering
   \includegraphics[width=1.0\linewidth]{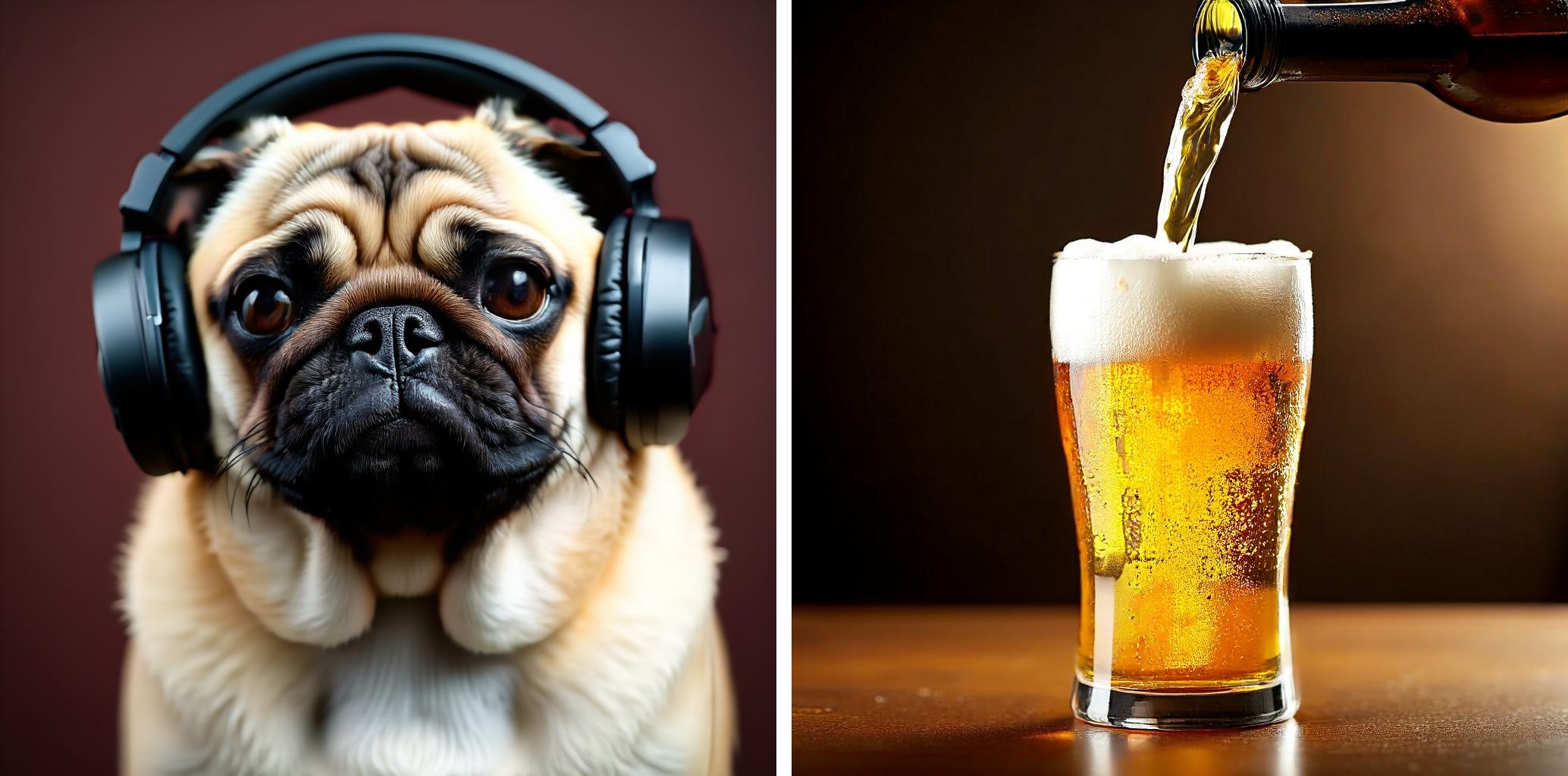}
   \caption{"Pug dog listening to music with big headphones.", "Low angle of pouring beer into a glass cup."}
\end{figure}

\begin{figure}[H]
  \centering
   \includegraphics[width=1.0\linewidth]{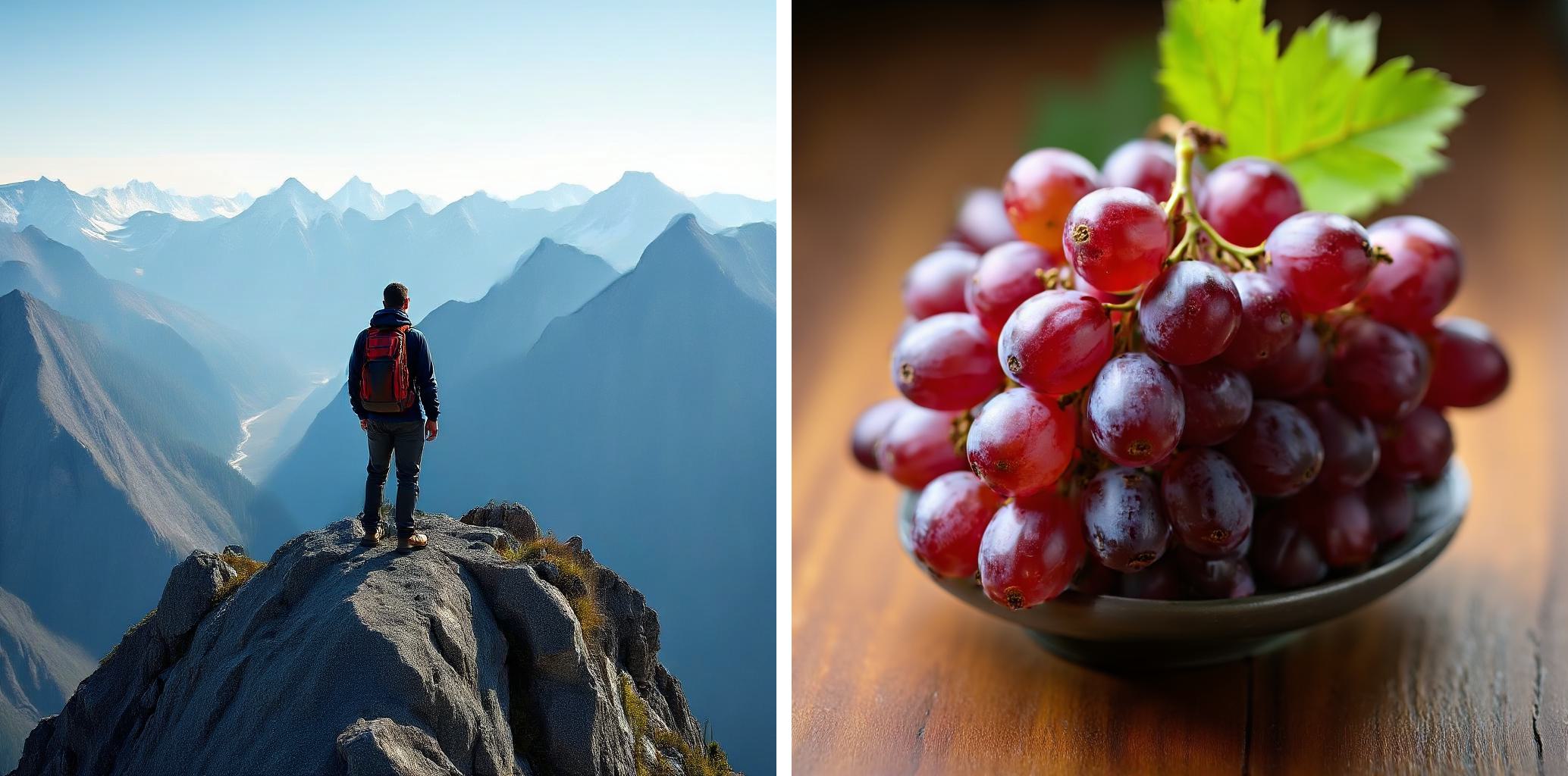}
   \caption{"Aerial view of a hiker man standing on a mountain peak.", "Close up of grapes on a rotating table. High definition."}
\end{figure}
\vspace{-10pt}

\begin{figure}[H]
  \centering
   \includegraphics[width=1.0\linewidth]{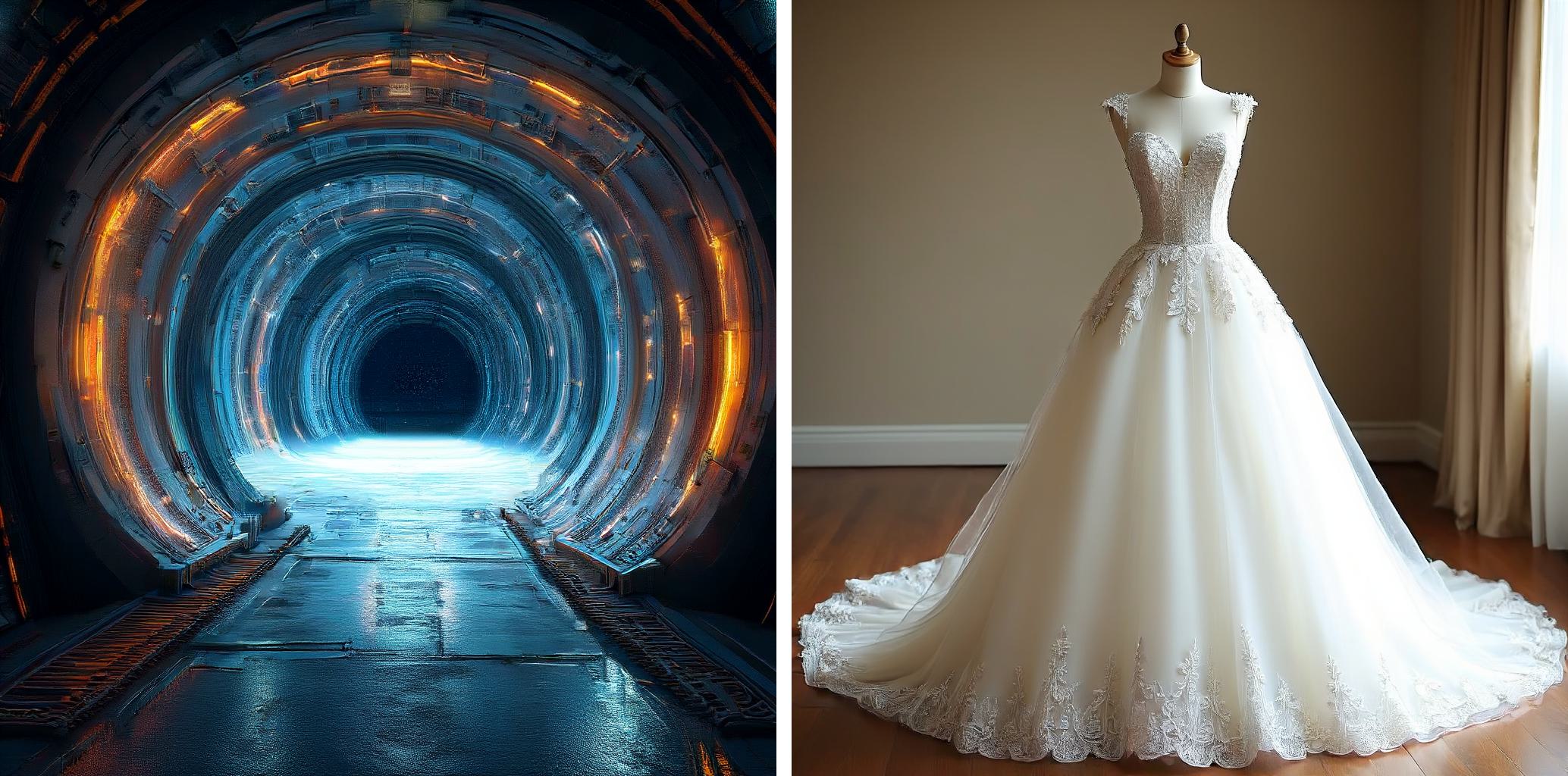}
   \caption{"Time and Space Tunnel", "Wedding dress"}
\end{figure}

\begin{figure}[H]
  \centering
   \includegraphics[width=1.0\linewidth]{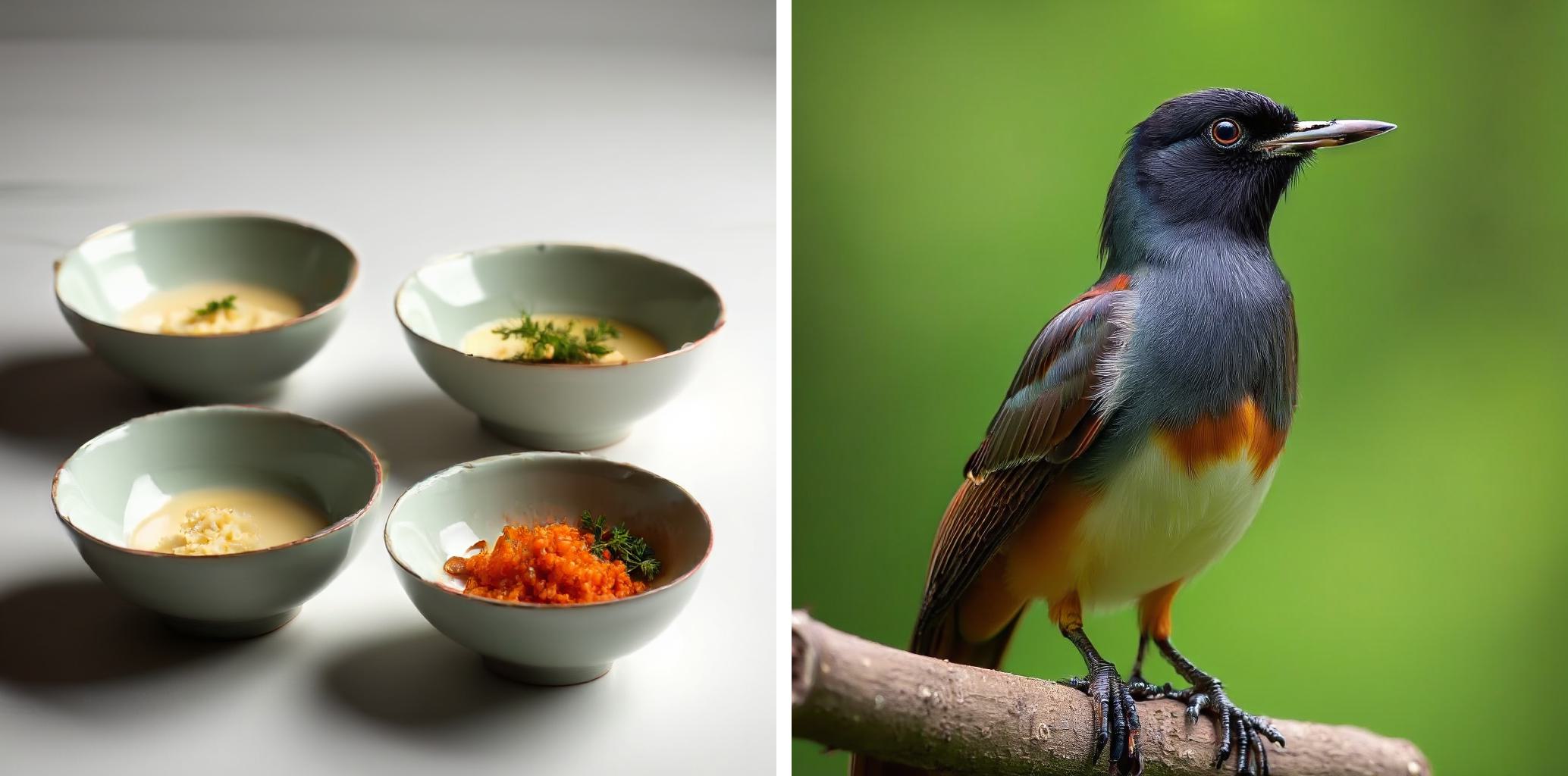}
   \caption{"a photo of four bowls", "a photo of a bird"}
\end{figure}

\begin{figure}[H]
  \centering
   \includegraphics[width=1.0\linewidth]{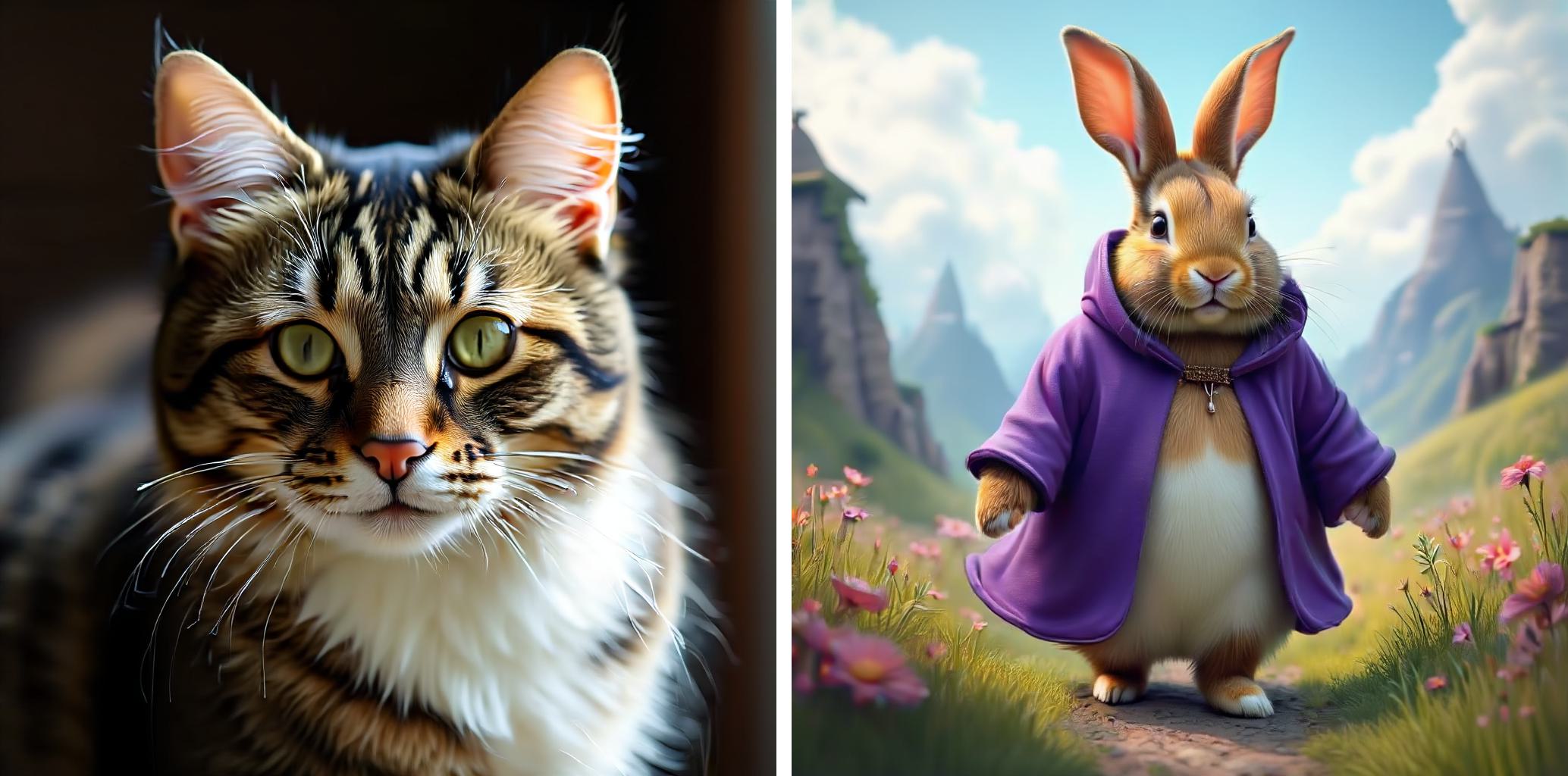}
   \caption{"a photo of a cat", "A fat rabbit wearing a purple robe walking through a fantasy landscape."}
\end{figure}

\begin{figure}[H]
  \centering
   \includegraphics[width=1.0\linewidth]{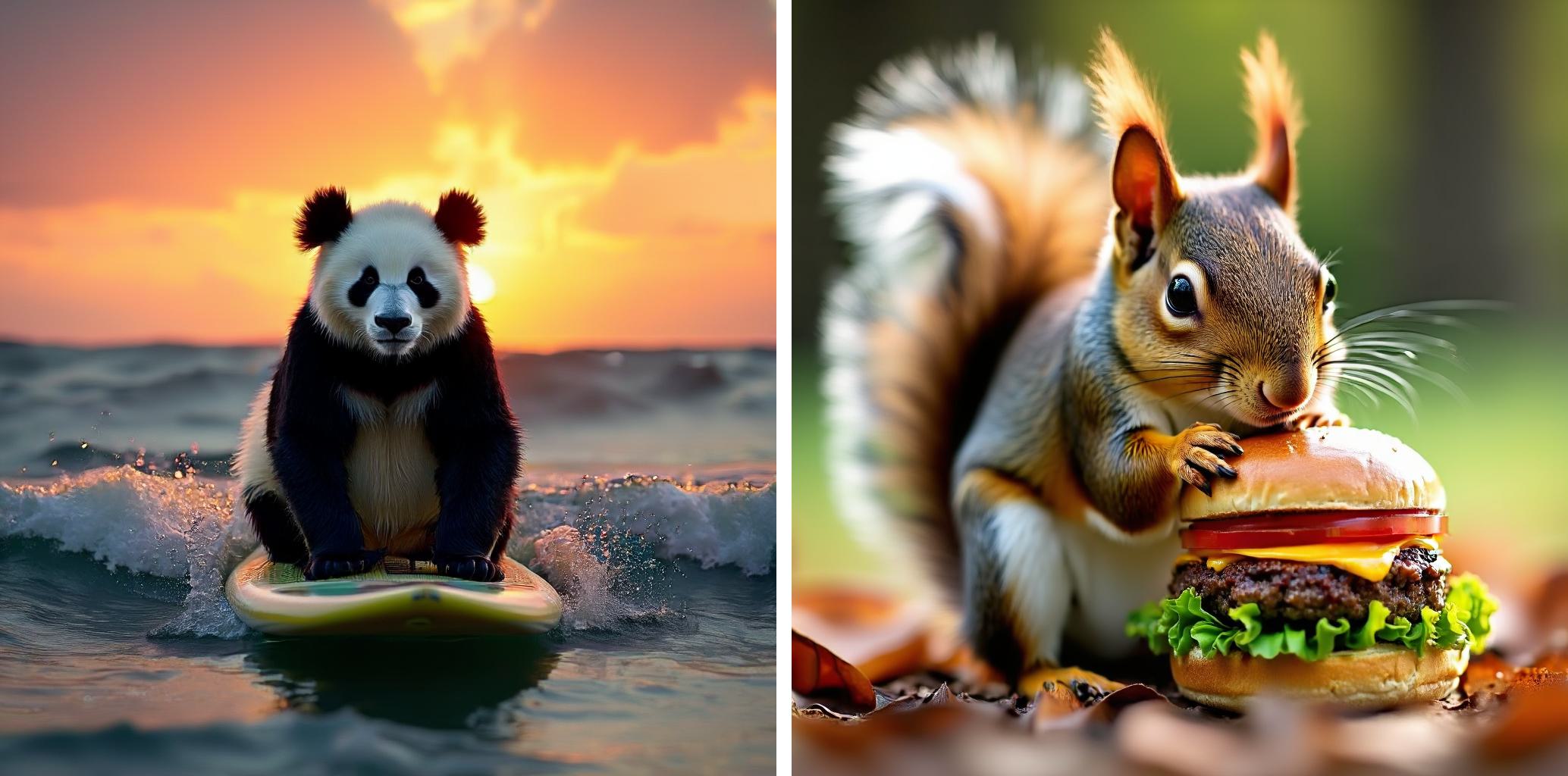}
   \caption{"A panda standing on a surfboard in the ocean in sunset, 4k, high resolution.", "A squirrel eating a burger."}
\end{figure}

\begin{figure}[H]
  \centering
   \includegraphics[width=1.0\linewidth]{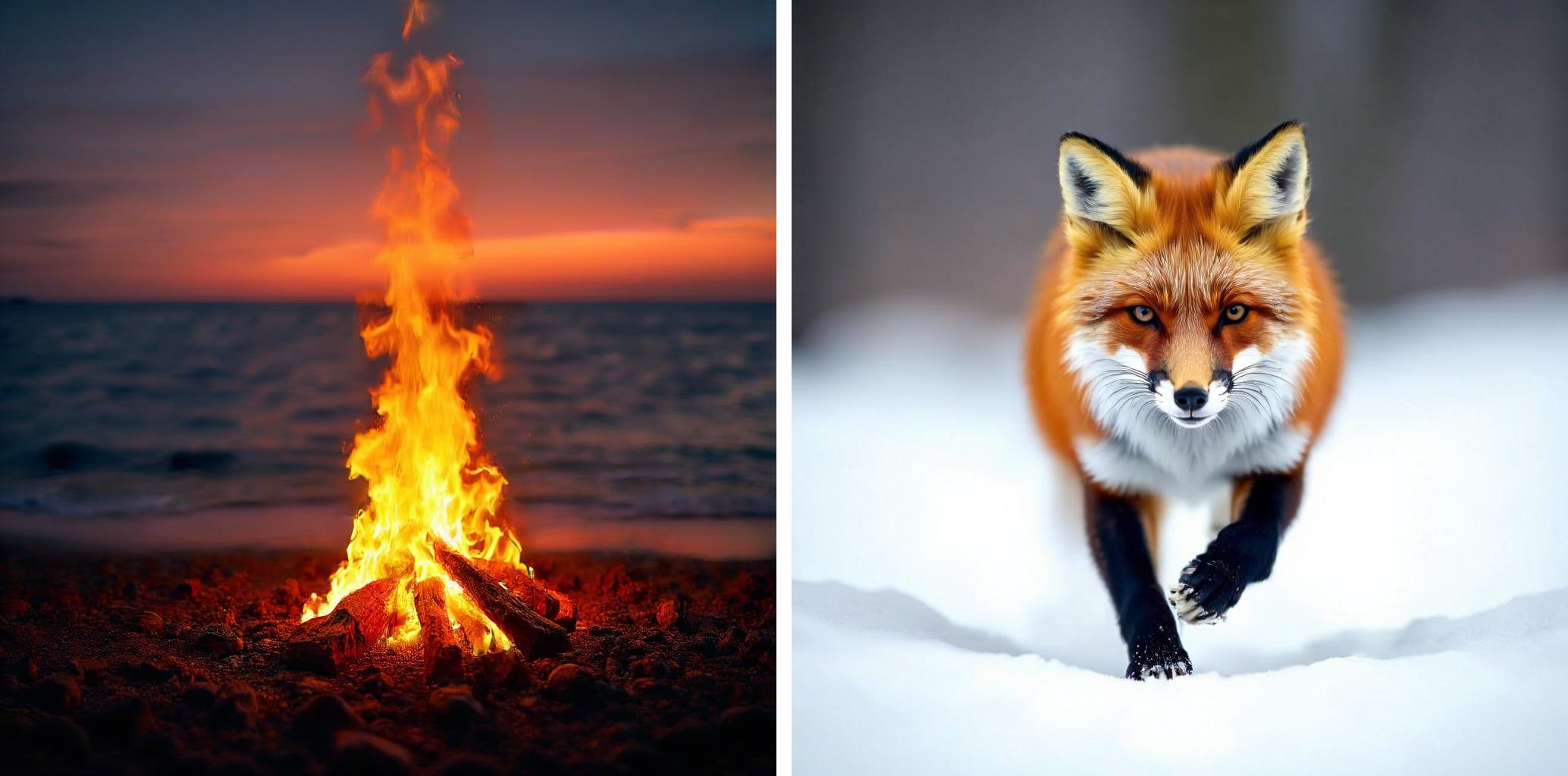}
   \caption{"Watch the fire from the shore", "red fox running in the snow"}
\end{figure}
\vspace{-10pt}

\begin{figure}[H]
  \centering
   \includegraphics[width=1.0\linewidth]{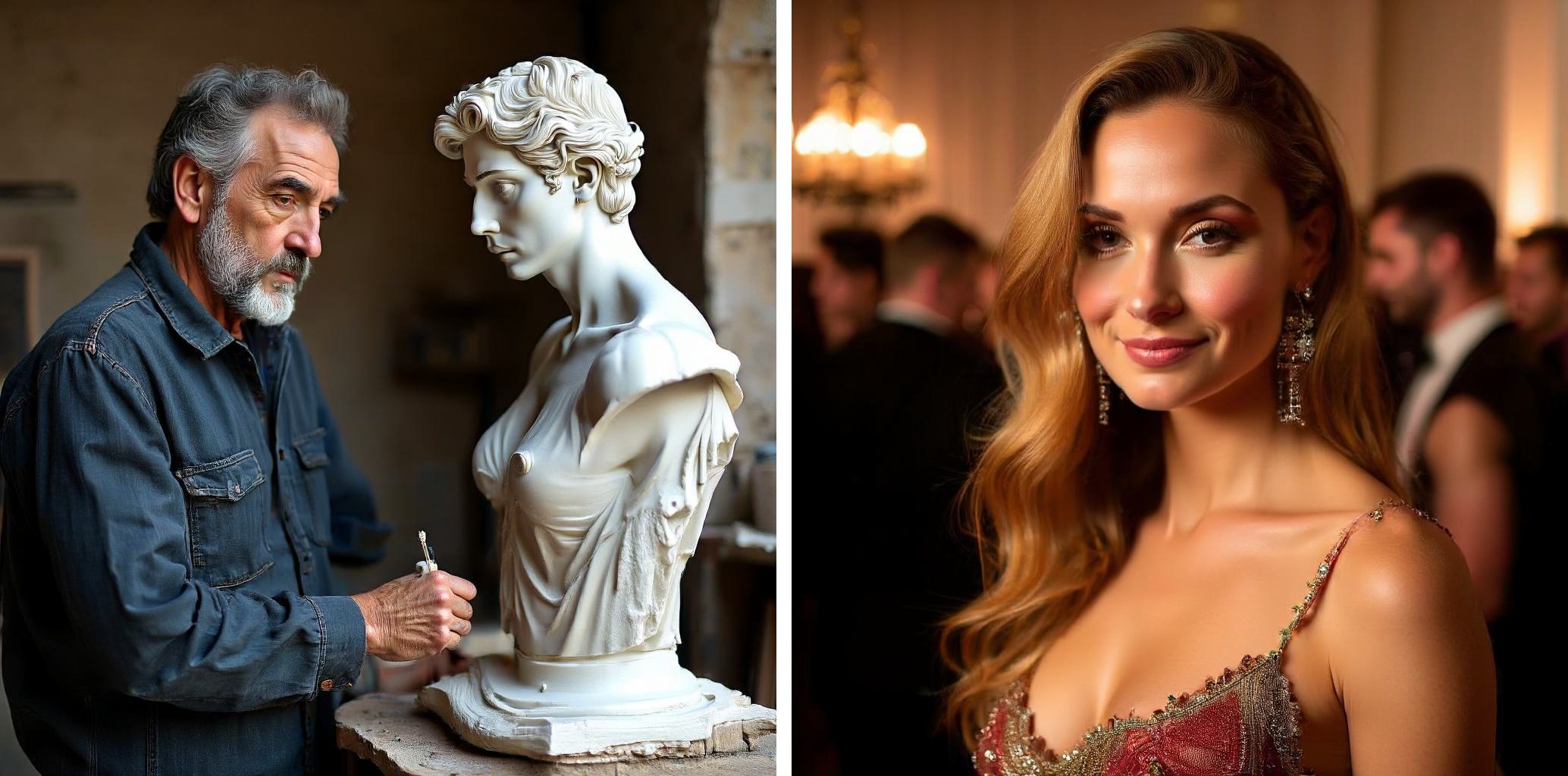}
   \caption{"Cesar, french sculptor, in his studio", "Grace Gummer at the party"}
\end{figure}

\begin{figure}[H]
  \centering
   \includegraphics[width=1.0\linewidth]{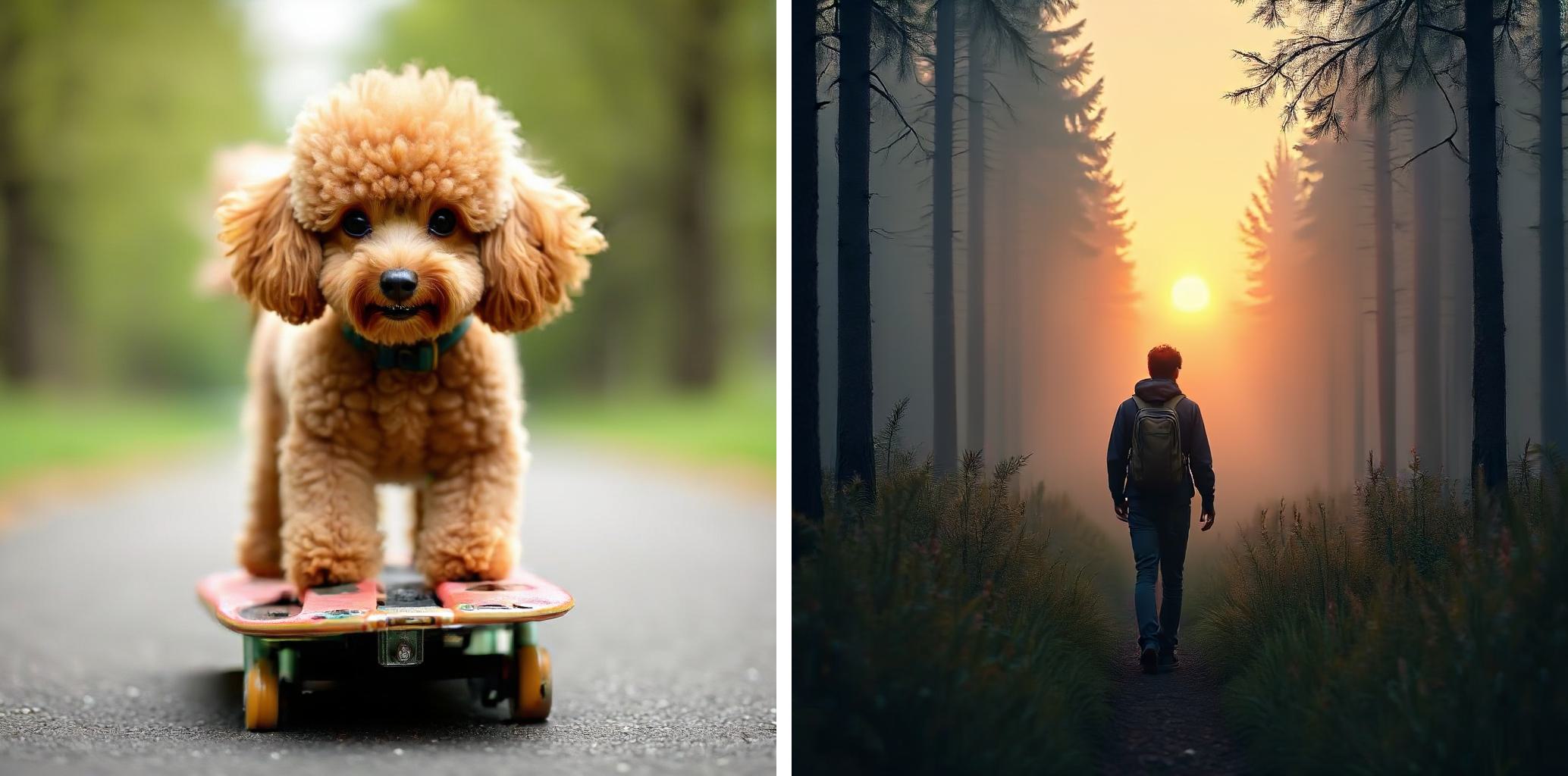}
   \caption{"Toy Poodle dog rides a penny board outdoors", "Traveler walking alone in the misty forest at sunset."}
\end{figure}

\vspace{-10pt}

\begin{figure}[H]
  \centering
   \includegraphics[width=1.0\linewidth]{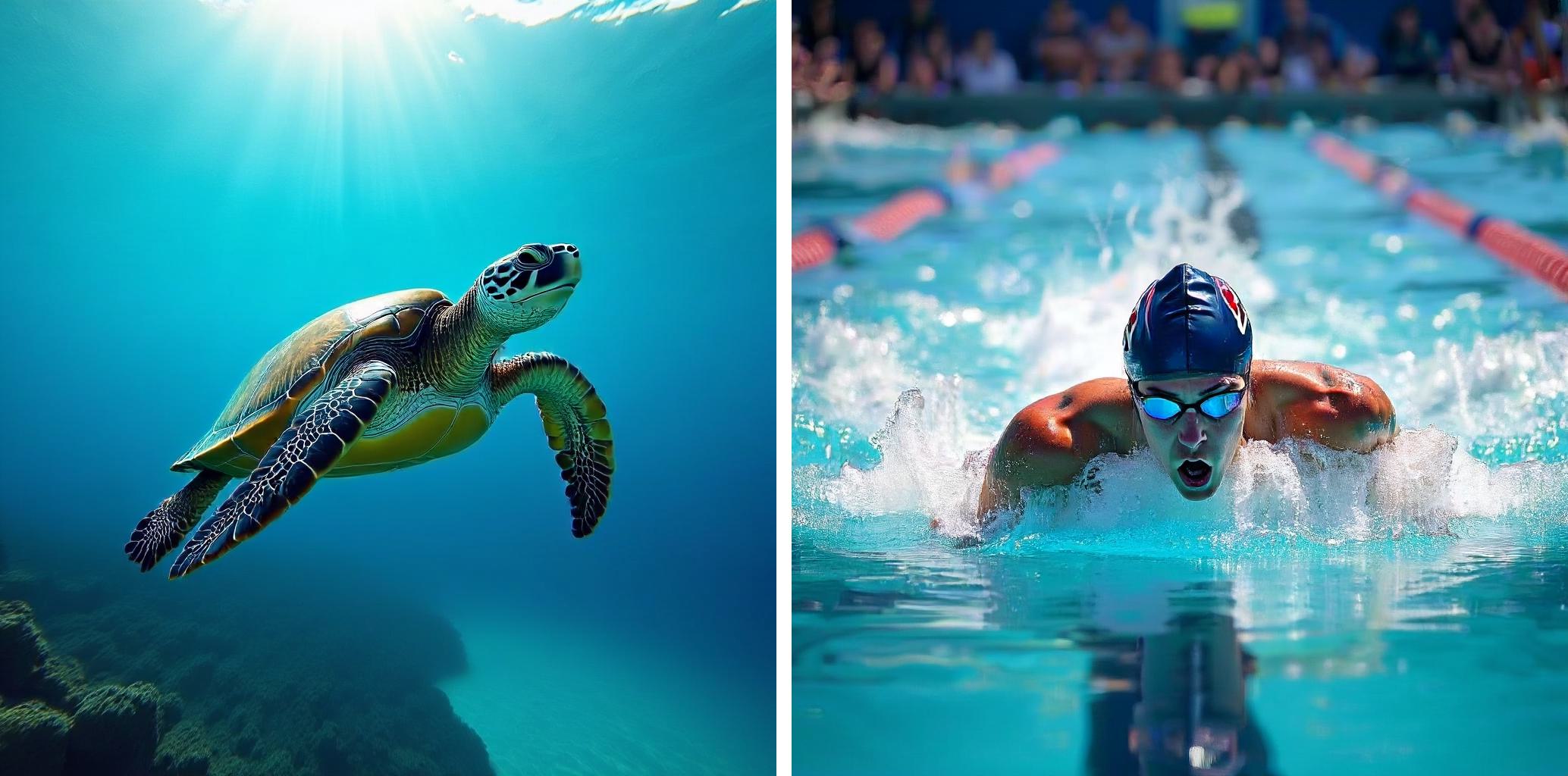}
   \caption{"Turtle swimming in ocean.", "PHOTOS: Life in the fast lane for our Olympic swimmers"}
\end{figure}

\begin{figure}[H]
  \centering
   \includegraphics[width=1.0\linewidth]{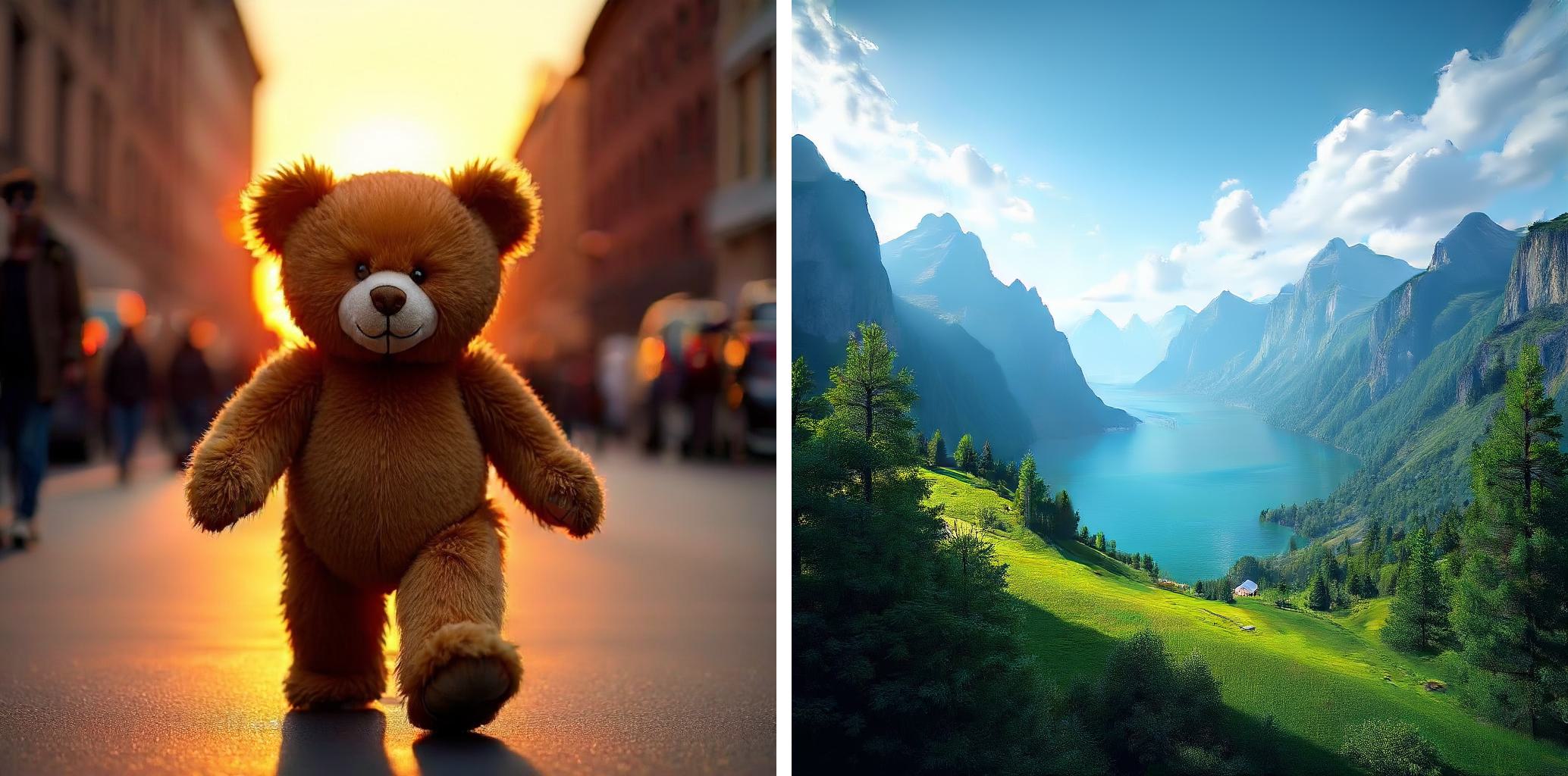}
   \caption{"Teddy bear walking down 5th Avenue, front view, beautiful sunset, close up, high definition, 4k.", "Time lapse at a fantasy landscape, 4k, high resolution."}
\end{figure}

\begin{figure}[H]
  \centering
   \includegraphics[width=1.0\linewidth]{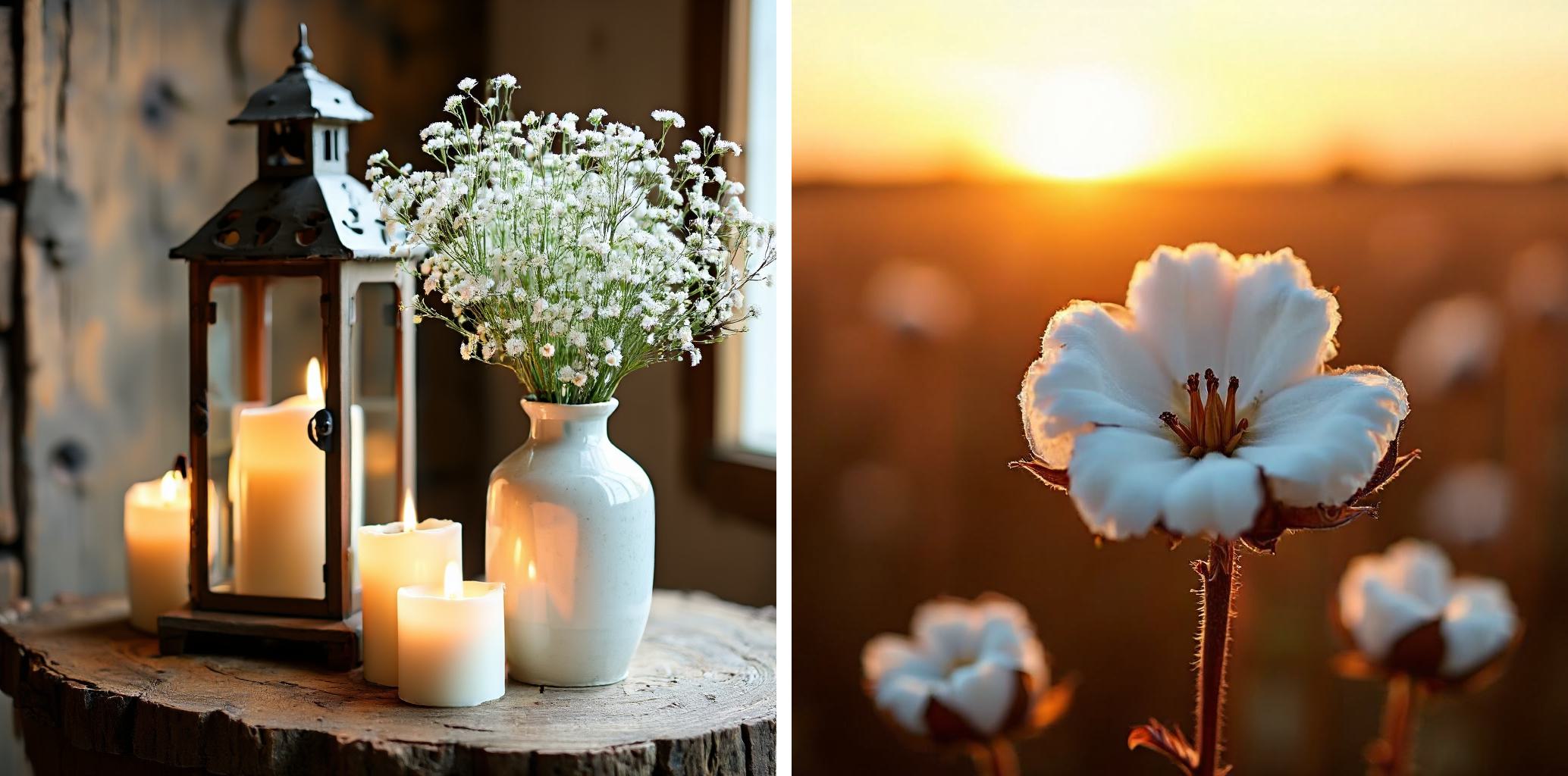}
   \caption{"a rustic lantern of wood, candles around and vintage vases with baby's breath", "cotton picking season. blooming cotton field. close up of the crop before the harvest, under a golden sunset light. - cotton stock videos \& royalty-free footage."}
\end{figure}

\begin{figure}[H]
  \centering
   \includegraphics[width=1.0\linewidth]{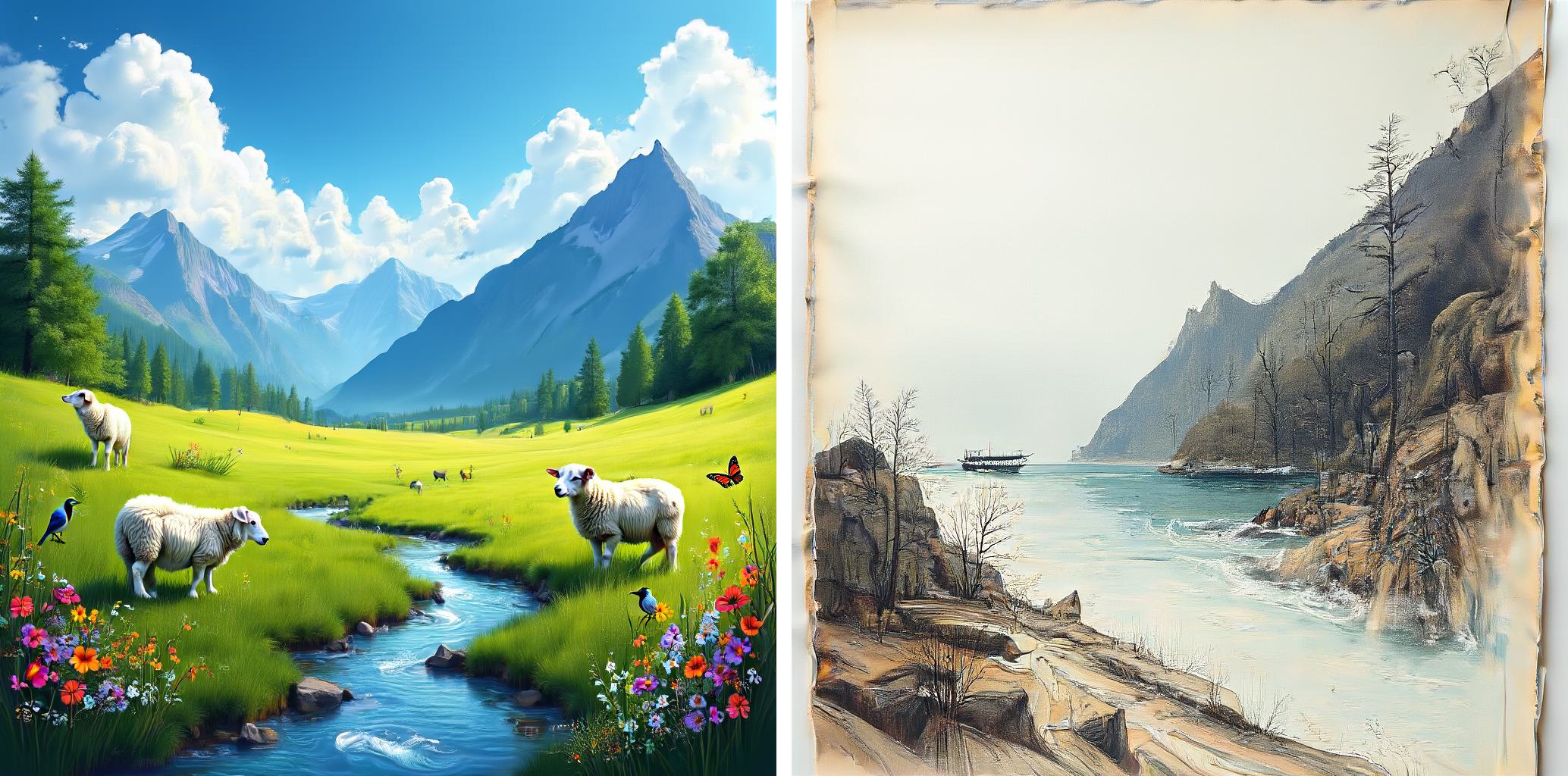}
   \caption{"a vast grassland appears particularly green under the sunlight, dotted with colorful wildflowers. A gentle breeze causes waves in the grass. In the distance, rolling mountains stand majestically against the backdrop of blue sky and white clouds. A few white sheep graze leisurely on the grass, with a shepherd dog guarding nearby. In the foreground, a clear stream meanders through, its water sparkling in the sunlight. Butterflies flutter among the flowers, and birds sing joyfully on the branches. The overall picture is rendered in a realistic style, with rich details and vibrant colors.", "Talk on paper"}
\end{figure}

\begin{figure}[H]
  \centering
   \includegraphics[width=1.0\linewidth]{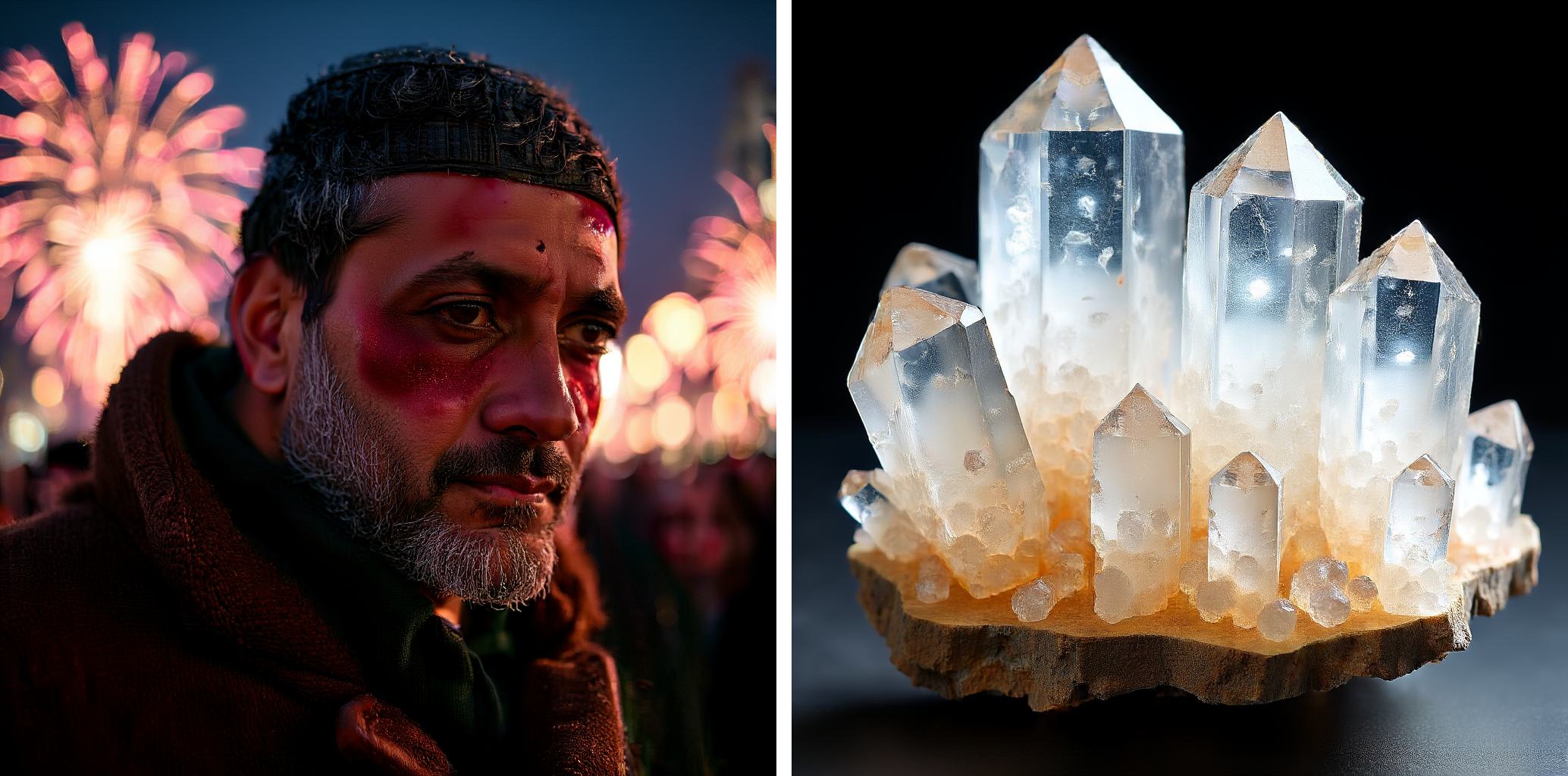}
   \caption{"Omar Elabdellaoui injured in the eye as a result of the accident at the New Year's Eve celebration", "A specimen of a variety of quartz showing conchoidal fracture"}
\end{figure}

\begin{figure}[H]
  \centering
   \includegraphics[width=1.0\linewidth]{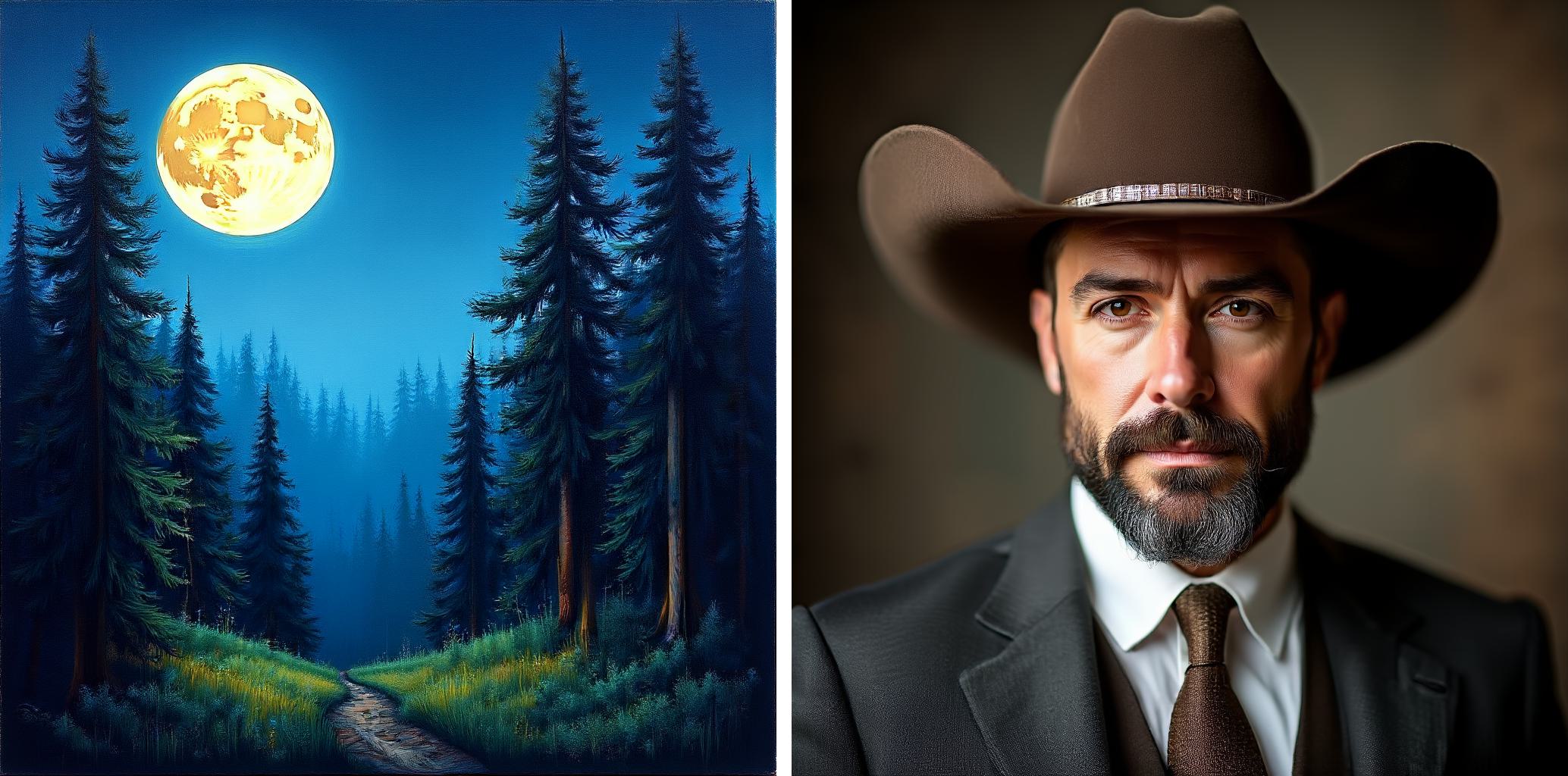}
   \caption{"The image features a beautiful painting of a forest at night, with a full moon illuminating the scene. The moon is positioned in the upper left corner of the painting, and its light casts a glow on the trees and the surrounding area. The forest is filled with trees, some of which are taller and closer to the foreground, while others are smaller and further away. The painting captures the serene atmosphere of the nighttime forest, with the moon as the main focal point.", "a man wearing a cowboy hat and a suit, with a beard and mustache. He is looking directly at the camera, giving the impression of a confident and distinguished appearance. The man's attire and hat suggest that he might be a cowboy or a businessman with a unique sense of style. The overall atmosphere of the image is one of sophistication and confidence."}
\end{figure}

\section*{\textbf{C.} More Discussion on the BridgeFlow Method}
The proposed Linear BridgeFlow ensures cross-stage velocity consistency. In contrast, nonlinear mappings tend to induce trajectory curvature and destabilize stage-wise optimization. Specifically, the conditional probability path is defined as:
\begin{equation}
\resizebox{0.7\linewidth}{!}{$
\hat{x}_{s_k} \sim \mathcal{N}\!\left(
\alpha_k \odot \mathrm{Up}\!\left(\mathrm{Down}\!\left(x_1\right)\right),
\Sigma_k
\right),
$}
\label{eq:xs}
\end{equation}
\begin{equation}
\resizebox{0.7\linewidth}{!}{$
\hat{x}_{e_{k-1}} \sim \mathcal{N}\!\left(
\beta_{k-1} \odot \mathrm{Down}\!\left(x_1\right),
\Sigma_{k-1}
\right),
$}
\label{eq:es}
\end{equation}
we first apply a simple upsampling transformation to achieve resolution alignment
\begin{equation}
\resizebox{0.8\linewidth}{!}{$
    \mathrm{Up}\!\big(\hat{x}_{e_{k-1}}\big) \sim \mathcal{N}\!\big(\beta^{'}_{k-1} \odot \mathrm{Up}\!\big(\mathrm{Down}\!\big(x_1\big)\big),
    \Sigma^{'}_{k-1}
\big).$}
\label{eq:Upes}
\end{equation}

We can further complete the distribution alignment by learning a linear transformation
\begin{equation}
    \hat{x}_{s_k} = W\cdot\mathrm{Up}\!\left(\hat{x}_{e_{k-1}}\right) + B
\label{eq:linear}
\end{equation}
to match the means and covariances of the two distributions.

We also provide a comparison of different bridging strategies in Table 7 of the main text, showing that more complex modules do not yield further improvements.

\section*{\textbf{D.} Example of NAMI for image editing}
\begin{figure}[H]
  \centering
 
  \includegraphics[width=1.0\linewidth]{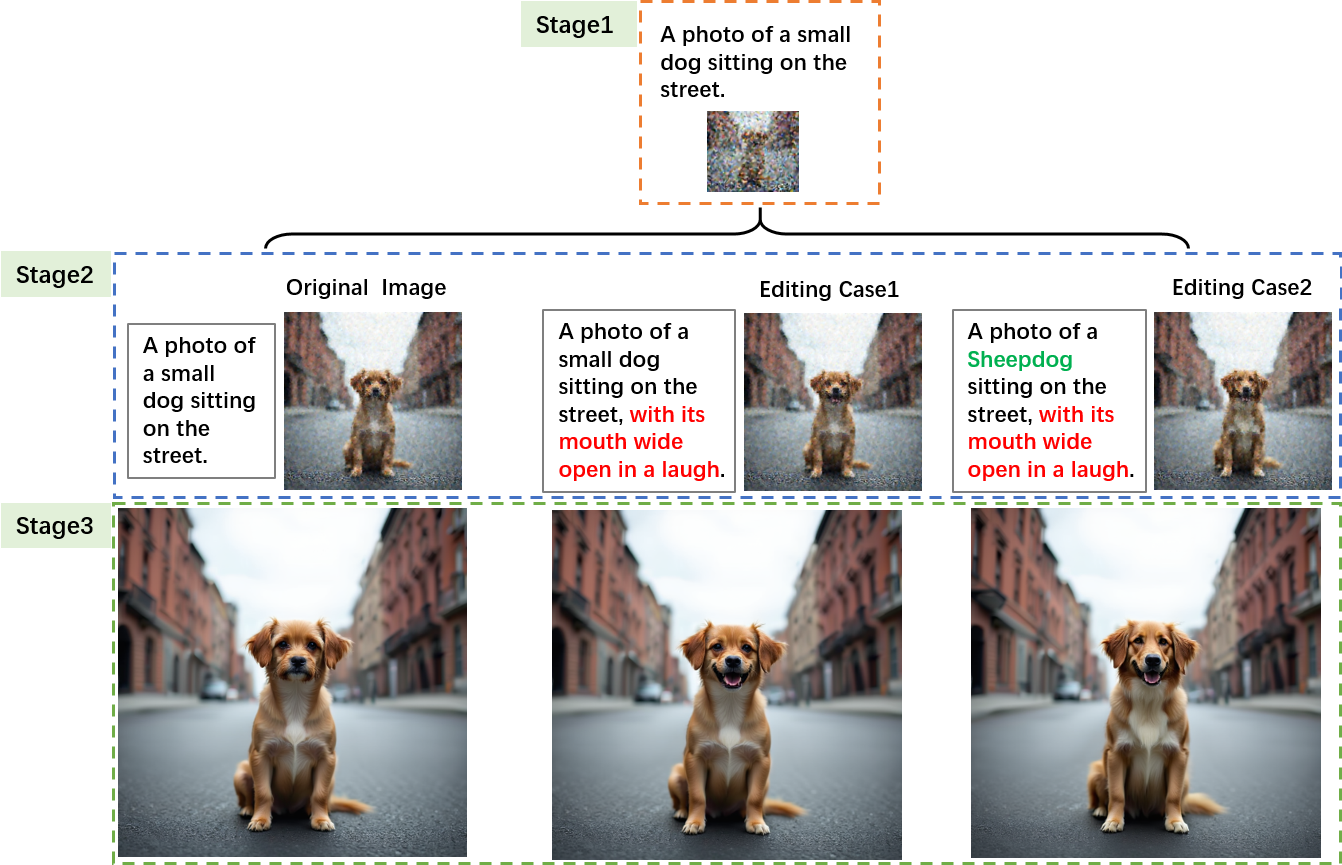}
  
   \caption{Image Editing of Directly Applying NAMI.}
   \label{fig:image edit}
\end{figure}

 We provide a simple, training-free example of directly applying NAMI for image editing. As shown in Figure \ref{fig:image edit}, we share the layouts and concept contours generated in the first stage, and modify the instructions in the subsequent stages to obtain edited images with different attributes. This demonstrates that NAMI is also convenient and promising for applications in other tasks.

\section*{\textbf{E.} Detailed explanation of the internal dataset}
Our internal dataset contains approximately 100,000 high-quality image-text pairs, curated for text-to-image generation tasks and primarily intended for high-quality fine-tuning of generative models. The data was collected from diverse sources across the internet, covering a wide range of topics and visual styles to ensure rich diversity in both content and semantics.
To improve caption quality and alignment, we used the CogVLM2 model to generate recaptions for each pair. The resulting image-text pairs then underwent multi-stage filtering as a whole, including evaluations of semantic coherence, aesthetic quality, and safety. Human assessments were also performed on the pairs to further ensure correctness, appropriateness, and overall quality. This rigorous process produced a meticulously curated dataset suitable for high-fidelity model fine-tuning.

\section*{\textbf{F.} The visualization of different methods}
\begin{figure*}
  \centering
  
   \includegraphics[width=1.0\linewidth]{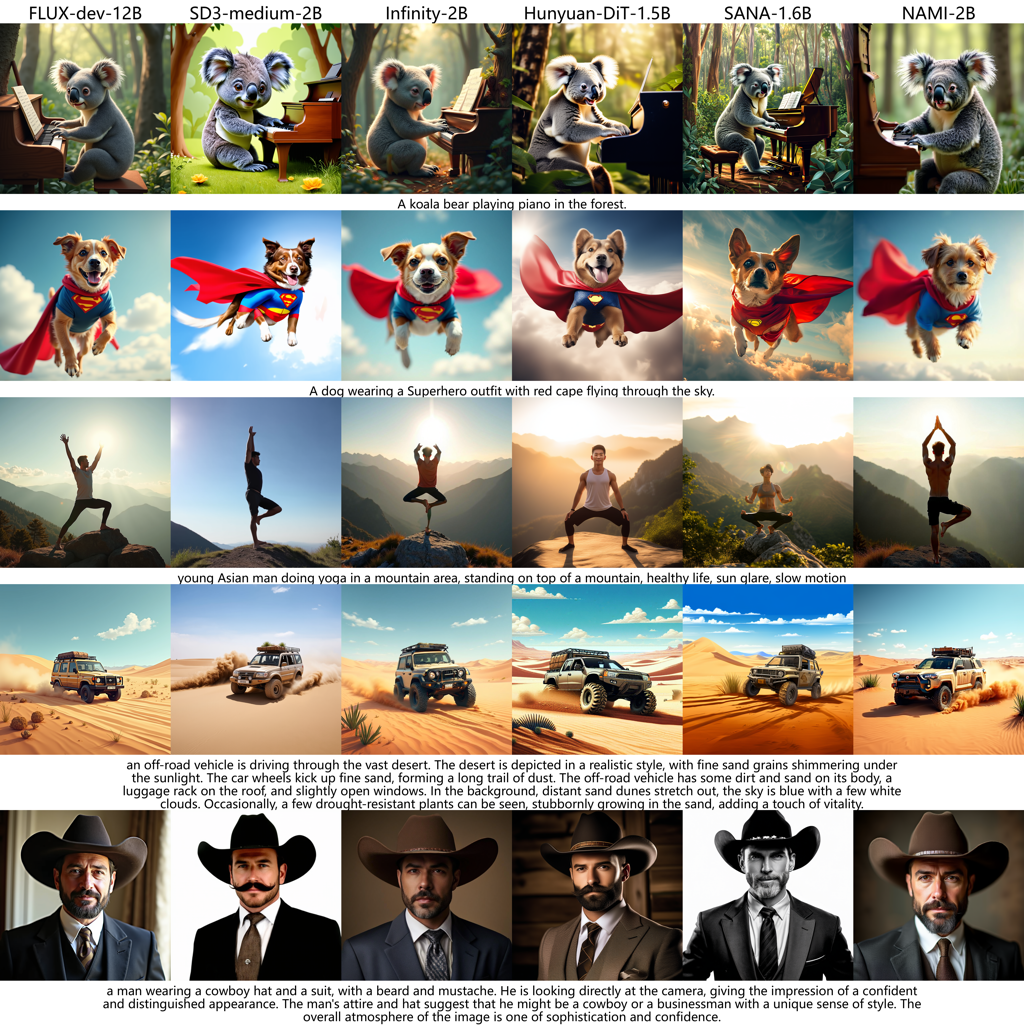}
   \caption{Visualization of image generation result comparison across different methods}
   \label{comparison across different methods}
\end{figure*}

\end{document}